\documentclass{article}
\usepackage{iclr2027_conference,times}

\usepackage{amsmath,amsfonts,bm}

\def\eqref#1{equation~\ref{#1}}
\def\1{\bm{1}}

\DeclareMathAlphabet{\mathsfit}{\encodingdefault}{\sfdefault}{m}{sl}
\SetMathAlphabet{\mathsfit}{bold}{\encodingdefault}{\sfdefault}{bx}{n}

\usepackage{caption}
\usepackage{hyperref}
\usepackage{url}
\usepackage{algorithm}
\usepackage{algpseudocode}
 \usepackage{microtype}

\usepackage[most]{tcolorbox}
\usepackage{listings}
\usepackage{multirow}
\usepackage{times}
\usepackage{latexsym}
\usepackage{amsmath, amssymb}
\usepackage{algorithm}

\usepackage[T1]{fontenc}
\usepackage{xcolor}

\definecolor{myred}{RGB}{192,0,0}
\definecolor{myblue}{RGB}{0,0,192}
\usepackage[utf8]{inputenc}

\usepackage{microtype}

\usepackage{inconsolata}
\newtheorem{proposition}{Proposition}
\usepackage{graphicx}
\usepackage{amsmath}
\usepackage{amssymb}
\usepackage{amsfonts}
\usepackage{booktabs}
\usepackage{geometry}
\usepackage[table]{xcolor}
\usepackage{booktabs}
\usepackage{wrapfig}
\title{From Rewarding What Models Say to How They Think: Attention Process Reward Model for LLM Reinforcement Learning}

\iclrfinalcopy

\author{
\textbf{
Kairong Han$^{1}$, Nuanqiao Shan$^{1}$, Ziyu Zhao$^{1}$,
Xinpeng Dong$^{1}$, Zijing Hu$^{1}$, Dian Jin$^{1}$,
}\\
\textbf{
Leyan Deng$^{2}$, Junjian Ye$^{2}$,
Lujia Pan$^{2}$, Fei Wu$^{1}$, Kun Kuang$^{1}$
}\\
$^{1}$Zhejiang University,
$^{2}$Noah's Ark Lab, Huawei Technologies\\
}

\begin{document}

\maketitle

\lhead{}
\begin{abstract}

Reinforcement learning (RL) has become an important component in large language models (LLMs). However, existing RL approaches focus on final outputs or textual solution trajectories, leading to coarse-grained rewards and unreliable reasoning. Such paradigms focus mainly on what the model outputs while overlooking how the model internally reasons. Therefore, can we reward how a model thinks internally rather than merely what it says? We first reveal the problem of textual reward aliasing in existing reward modeling approaches. Motivated by this, we propose the \underline{\textbf{A}}ttention \underline{\textbf{P}}rocess \underline{\textbf{R}}eward \underline{\textbf{M}}odel (APRM), which assigns internal rewards based on hidden states and attention maps to capture high-quality reasoning patterns. With only approximately 46M parameters, APRM is trained using ranking and counterfactual objectives, and its reward guides RL optimization while being periodically updated to handle distribution shifts. Integrated with representative RL methods, APRM achieved an average pass@8 gain of 2.88\%. Our results demonstrate the promise of incorporating internal reasoning signals into RL for LLMs. 
\end{abstract}

\section{Introduction}

\begin{wrapfigure}{r}{0.58\columnwidth}
    \vspace{-15pt}
    \centering
    \includegraphics[width=0.58\columnwidth]{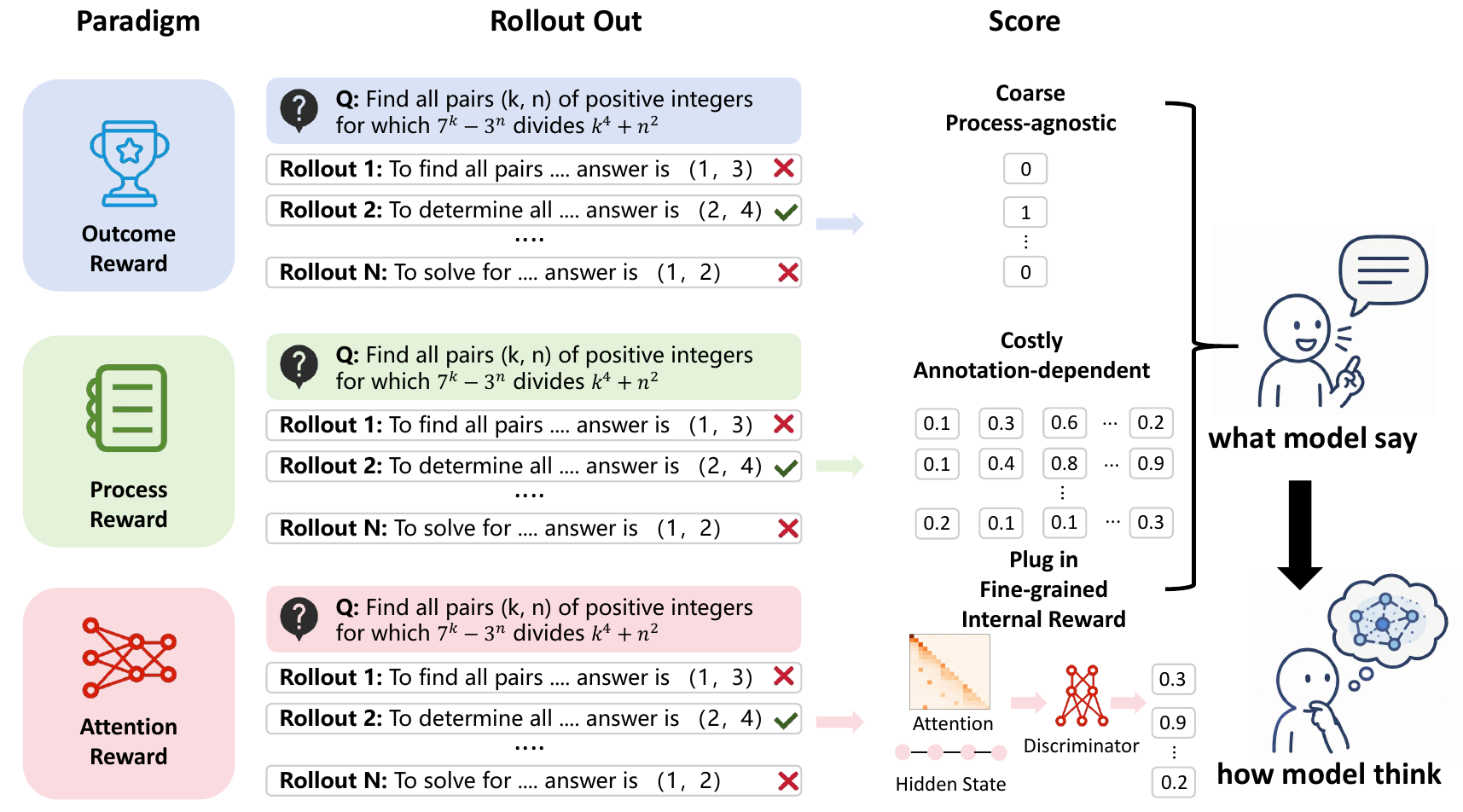}
    \caption{Comparison of outcome rewards, textual process rewards, and APRM.}
    \label{fig:Intro}
    \vspace{-10pt}
\end{wrapfigure}
In recent years, large language models (LLMs) \citep{zhao2023survey,chang2024survey,vaswani2017attention} have demonstrated remarkable capabilities in complex reasoning tasks \citep{wang2024survey,plaat2025agentic}. Reinforcement learning (RL) has consequently become an important approach for improving reasoning ability during the post-training stage \citep{cao2024survey,zhang2025survey}.

Existing RL approaches for reasoning typically employ outcome reward models (ORMs) \citep{zhong2025comprehensive} and process reward models (PRMs) \citep{xi2026agentprm,gao2024designing}. ORMs reward final-answer correctness, which is scalable but provides only sparse outcome-level feedback and cannot identify which reasoning steps contribute to success or failure. PRMs provide denser supervision by scoring intermediate Chain-of-Thought (CoT) steps, but they rely on natural-language rationales that are costly to annotate.  Moreover, existing reward models supervise only externalized reasoning traces. However, reasoning in LLMs is fundamentally latent \citep{chen2025reasoning,barez2025chain}. The internal computation that produces a textual trajectory is not directly observable from the trajectory itself.  Therefore, textual reward supervision may suffer from process aliasing,
where distinct internal reasoning processes induce indistinguishable textual outputs and consequently receive identical rewards. Therefore, a question arises:

\textit{Can we reward how a model thinks internally, rather than merely what it says?}

Mechanistic interpretability studies reveal structured information routing and interpretable computations in attention components~\citep{lan2024towards,merullo2024talking,kissane2024interpreting,nam2026causal}, so we use attention-derived signals as a partial window into internal computation, extending reward modeling from \textit{what models say} toward \textit{how they think}. We first theoretically show that textual reward supervision is fundamentally under-observed. Distinct internal reasoning processes may collapse to indistinguishable textual trajectories and consequently receive identical rewards, a phenomenon we term \textbf{textual reward aliasing}. Motivated by this view, we propose the Attention Process Reward Model (APRM), as shown in Figure \ref{fig:Intro}. The APRM is built upon an Attention-Fused Concept Pooler and an Edge-Biased Graph Attention Network (GAT) \citep{velivckovic2017graph}. APRM directly evaluates hidden states and attention maps from the reasoning process to produce internal process rewards. We use APRM in two stages. During cold-start, APRM uses positive and negative reasoning trajectories with ranking and counterfactual losses to learn preferences over internal reasoning processes. During RL, APRM reshapes the task reward so that policy optimization is guided by both outcomes and internal attention quality.

Through systematic analysis of APRM, we obtain three key findings: (1) Positive and negative trajectories exhibit systematic structural differences in the APRM concept-attention space. (2) APRM captures meaningful attention-derived signals and is sensitive to counterfactual attention perturbations. (3) APRM is not a proxy for existing metrics. It exhibits length-decoupling behavior and captures complementary process-quality information beyond existing reward supervision.

We further validate APRM by integrating it with representative RL methods. As a lightweight plug-in module with only approximately 46M parameters, APRM consistently improves avg@8 and pass@8, yielding an average pass@8 gain of 2.88\%. Interestingly, APRM also encourages more efficient reasoning behaviors, with a tendency toward shorter reasoning trajectories while maintaining or improving performance. Our contributions are summarized as follows:
\begin{itemize}

    \item We reveal the phenomenon of the textual reward aliasing problem and demonstrate the information gain of using internal signals as rewards.

    \item We propose APRM, a lightweight reward model with approximately 46M parameters that evaluates reasoning trajectories using attention-derived internal signals.

    \item We systematically analyze the properties of APRM and further validate its effectiveness on downstream reinforcement learning tasks.

\end{itemize}

\section{Motivation and Theoretical View of APRM}

\paragraph{Problem Formulation.} Given a prompt $x \sim \mathcal{D}$, an autoregressive policy
$\pi_\theta$ generates a reasoning trajectory
$y \sim \pi_\theta(\cdot \mid x)$.
A verifier $V$ evaluates the final answer and provides an
outcome reward $r^{\mathrm{out}}=V(x,y)\in\{0,1\}$.
Meanwhile, the model produces internal representations,
including hidden states $\mathbf{H}$ and attention maps
$\mathbf{A}$.
Our objective is to improve reasoning performance by exploiting
both external outcome feedback and internal process information.
To this end, APRM, parameterized by $\phi$, estimates an internal
process reward $s^{\mathrm{APRM}}=S_\phi(\mathbf{H},\mathbf{A})$,
which is incorporated into policy optimization as an auxiliary
learning signal.

\begin{proposition}\label{proposition 1}
\textbf{(Textual Reward Aliasing)}.
In self-attention, representations are computed as $\mathbf{C} = \mathbf{A}\mathbf{V}$, where $\mathbf{A}$ is the attention matrix and $\mathbf{V}$ is the value matrix. When
$
\operatorname{rank}([\mathbf{V},\mathbf{1}]) < T,
$
where $T$ is the sequence length. There exist distinct attention matrices $\mathbf{A} \neq \mathbf{A}'$ such that
$
\mathbf{A}\mathbf{V} = \mathbf{A}'\mathbf{V}
$.
\end{proposition}

Proposition ~\ref{proposition 1} implies that distinct internal reasoning processes may induce indistinguishable textual trajectories and therefore receive identical rewards under outcome-level or textual supervision.
In long-chain CoT scenarios, $T$ typically far exceeds $V$. We refer to this phenomenon as \textbf{textual reward aliasing}. Existing ORMs and textual PRMs only observe external behavioral signals $B=b(X,Y)$, where $X$ denotes the input prompt and $Y$ denotes the generated trajectory. In contrast, APRM additionally incorporates latent reasoning states $Z=(\mathbf{H},\mathbf{A})$, consisting of hidden states $\mathbf{H}$ and attention dynamics $\mathbf{A}$, thereby expanding the observable space of reward modeling.  Full proofs are provided in Appendix~\ref{APP:Proof}.

\begin{proposition}\label{proposition 2}
\textbf{(Non-redundant Internal Supervision)}.
Let Q denote process quality. Under log-loss, APRM provides potential non-redundant information gain:
$
I(Q;Z\mid X,B)>0.
$
\end{proposition}

Proposition~\ref{proposition 2} suggests that internal attention structures contain process-quality information beyond textual outputs and existing reward signals. We empirically validate this condition in Section~\ref{Information Gain Validation} through a residual information test. Full proofs are provided in Appendix~\ref{APP:Proof}.

Proposition ~\ref{proposition 1} shows the existence of textual reward aliasing, and Proposition ~\ref{proposition 2} further shows that the lost information contains reasoning-relevant signals. Based on these findings, we propose APRM to capture this previously overlooked yet informative signal.

\section{Methods}

\subsection{APRM Architecture}

\begin{wrapfigure}{r}{0.48\columnwidth}
    \vspace{-15pt}
    \centering
\includegraphics[width=1\linewidth]{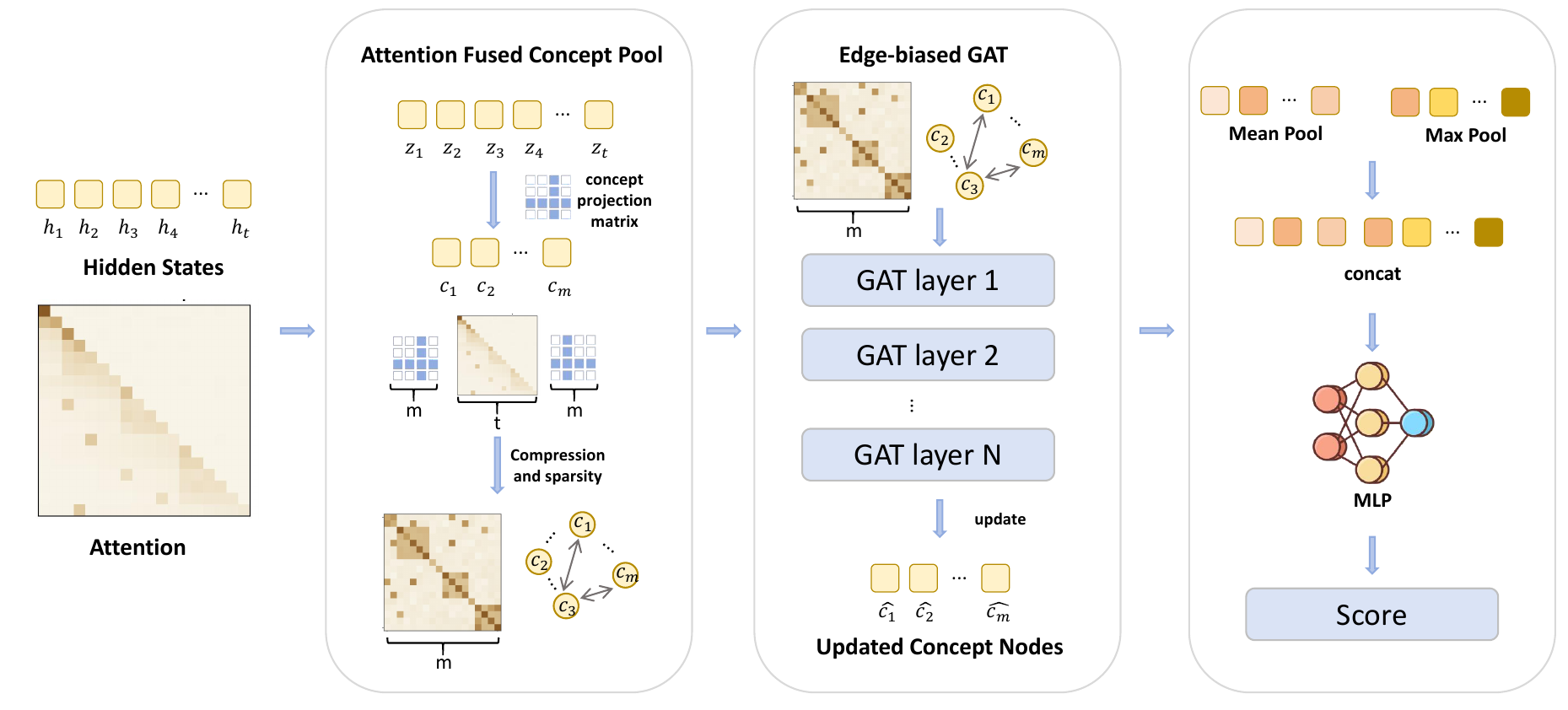}
    \caption{Architecture of APRM.}
    \label{fig:arm_arch}
    \vspace{-15pt}
\end{wrapfigure}
As shown in Figure~\ref{fig:arm_arch}, we propose APRM as an internal process reward model for Transformer-based LLMs. Given the hidden states and attention maps produced during generation, APRM estimates a scalar score that reflects the quality of internal representations.

\subsubsection{Attention-Fused Concept Pooling}

Reasoning structure is often distributed across tokens and attention interactions. APRM therefore compresses them into latent reasoning concepts, enabling structured process-level modeling while reducing memory overhead. For each rollout sample, APRM uses the final-layer hidden states
$\mathbf{H}\in\mathbb{R}^{T\times d}$ and constructs an average attention over heads and layers from the last $K$ layers:
\[
\mathbf{A}
=
\frac{1}{K}
\sum_{\ell=L-K+1}^{L}
\left(
\frac{1}{N_h}
\sum_{h=1}^{N_h}
\mathbf{A}^{(\ell,h)}
\right) \in\mathbb{R}^{T\times T},
\]
where $N_h$ denotes the number of attention heads.

Then we construct attention-aware token representations by fusing each token representation with its attention-weighted context:
\[
\begin{gathered}
\mathbf{Z}
=
\mathrm{LayerNorm}
\big(
\mathbf{H}
+
\mathrm{MLP}([\mathbf{H};\mathbf{A}\mathbf{H})
\big).
\end{gathered}
\]
To obtain latent reasoning concepts, APRM first uses a learnable projection matrix $\mathbf{W}c\in\mathbb{R}^{d\times M}$ to compute concept-assignment logits and applies softmax over the token dimension, followed by weighted pooling to derive the concept representations:
\[
\mathbf{W}=\mathrm{Softmax}{T}\left((\mathbf{Z}\mathbf{W}_c)^{\top}\right)\in\mathbb{R}^{M\times T},\qquad
\mathbf{C}=\mathbf{W}\mathbf{Z}\in\mathbb{R}^{M\times d}.
\]

Based on the same assignment matrix, APRM projects the token-level attention summary into the concept space and constructs a sparse normalized concept graph:
\[
\hat{\mathbf{A}}_c
=
\mathrm{TopKNorm}
\left(
\mathbf{W}\mathbf{A}\mathbf{W}^{\top}
\right)
\in
\mathbb{R}^{M\times M},
\]
where $\mathrm{TopKNorm}(\cdot)$ denotes row-wise top-$k$ sparsification followed by row normalization.

\subsubsection{Edge-Biased Graph Attention Network}

To enhance APRM's ability to represent the model's internal reasoning process and better capture the interactions between attention patterns and hidden states, we propose the Edge-Biased Graph Attention Network. Given the concept representations $\mathbf{C}$ and the sparse concept graph $\hat{\mathbf{A}}_c$, APRM applies stacked Edge-Biased GAT layers to model dependencies among latent reasoning concepts. For each concept edge, $\hat{\mathbf{A}}_{c,ij}$ is used both as an additive bias in the attention logits and as a multiplicative gate on the transmitted message:
\[
\begin{gathered}
\alpha_{ij}
=
\mathrm{Softmax}_{j}
\left(
\frac{\mathbf{Q}_i\mathbf{K}_j^{\top}}{\sqrt{d_h}}
+
b(\hat{\mathbf{A}}_{c,ij})
\right),
\mathbf{m}_{ij}
=
\alpha_{ij}
\,g(\hat{\mathbf{A}}_{c,ij})
\,\mathbf{V}_j .
\end{gathered}
\]
Here, $b(\cdot)$ and $g(\cdot)$ denote the edge bias and edge gate functions, respectively.

After graph propagation, APRM applies mean pooling and max pooling over concept nodes. The pooled representations are concatenated and fed into an MLP to produce the final internal process score:
\[
s_{\mathrm{APRM}}
=
\mathcal{S}_{\phi}(\mathbf{H},\mathbf{A}).
\]
Due to space limitations, detailed layer-wise operations are provided in Appendix~\ref{APP:Edge-Bias GAT}.
\begin{figure*}[t]
    \centering
    \includegraphics[width=0.8\linewidth]{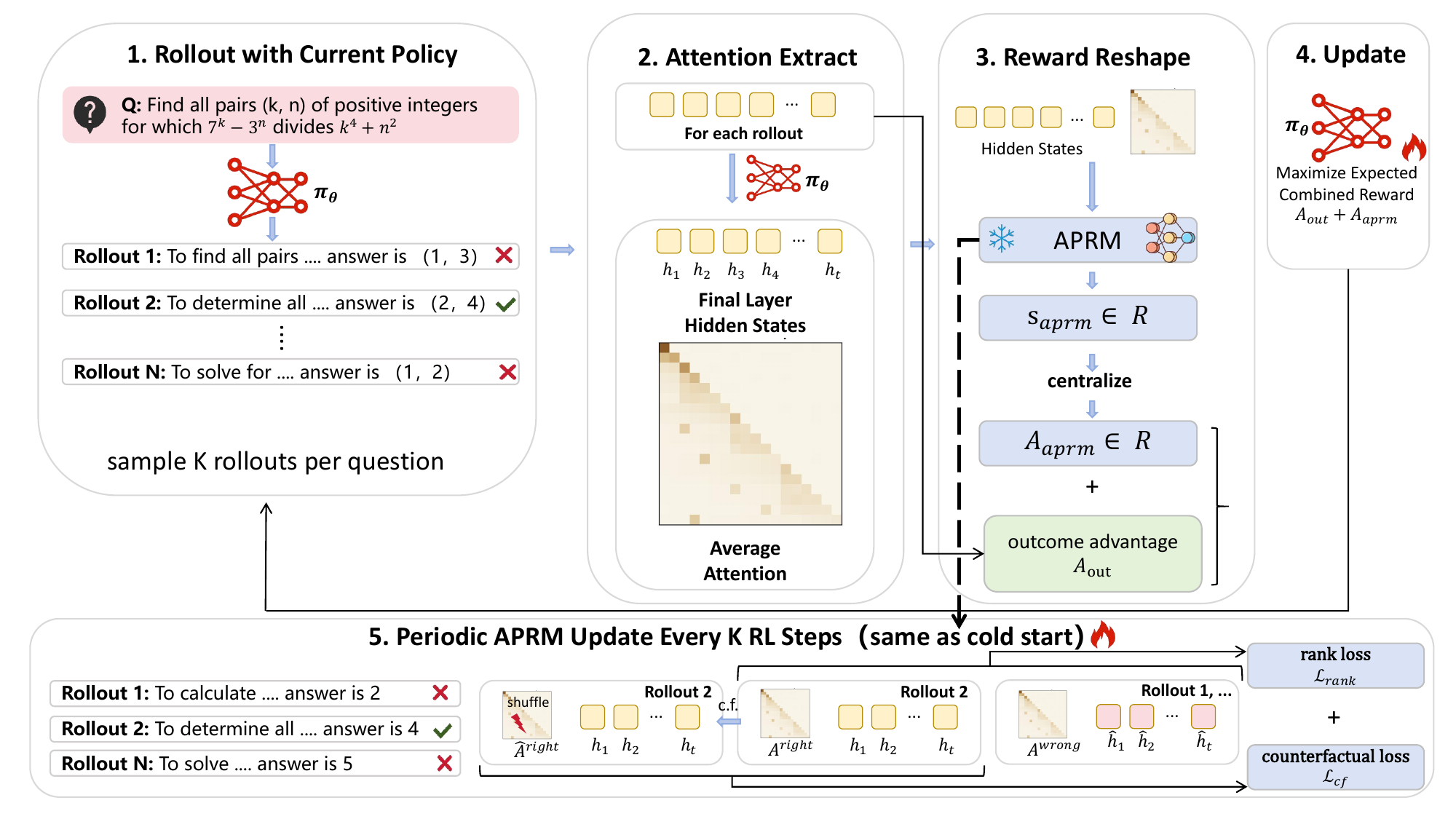}
    \caption{APRM-enhanced reinforcement learning workflow.APRM computes an internal process advantage from its reward scores and combines it with the outcome-based advantage, while being periodically refreshed to mitigate distribution shift.}
    \label{fig:process}
    \vspace{-1em}
\end{figure*}

\subsection{Cold-start Training of APRM}

To train APRM, we first construct a rollout dataset using the backbone LLM. For each question $x$, the model samples multiple reasoning trajectories: $\{y_i\}_{i=1}^{N}
\sim
\pi_{\theta_0}(\cdot\mid x),$ where sampling is performed with temperature and top-$p$ sampling. Each trajectory is labeled by an answer verifier: $r_i^{\mathrm{out}}\in\{0,1\}$,
where $r_i^{\mathrm{out}}=1$ indicates that the trajectory reaches the correct final answer. For each trajectory, we extract internal features $(\mathbf{H}_i,\mathbf{A}_i)$ from the backbone LLM and use them to train APRM. Thus, the APRM training dataset consists of tuples
$(x,y_i,\mathbf{H}_i,\mathbf{A}_i,r_i^{\mathrm{out}})$.

\textbf{Ranking Loss.} Given a correct trajectory $y^+$ and an incorrect trajectory $y^-$ for the same question, APRM is trained to assign a higher internal score to the correct trajectory. The pairwise ranking loss is
\[
\mathcal{L}_{\mathrm{rank}}
=
\max
\big(
0,
m_{rk}-(s^+-s^-)
\big),
\]
where $s^+=
\mathcal{S}_{\phi}(\mathbf{H}^+,\mathbf{A}^+)$, $
s^-=
\mathcal{S}_{\phi}(\mathbf{H}^-,\mathbf{A}^-)
$,
and $m_{rk}$ is the ranking margin.

\textbf{Counterfactual Loss.} To encourage APRM to capture informative attention structures, we construct counterfactual attention patterns from positive trajectories. Specifically, we perturb the original attention map using token-level dropout, token permutation, and Gaussian noise. Given the original attention map $\mathbf{A}$ and the perturbed counterpart $\tilde{\mathbf{A}}$, APRM is encouraged to assign a higher score to the original internal process:
\[
\mathcal{L}_{\mathrm{cf}}
=
\max
\left(
0,
m_{\mathrm{cf}}
-
\left[
\mathcal{S}_{\phi}(\mathbf{H},\mathbf{A})
-
\mathcal{S}_{\phi}(\mathbf{H},\tilde{\mathbf{A}})
\right]
\right),
\]
where $m_{\mathrm{cf}}$ is the counterfactual margin.

Finally, the overall cold-start objective is
\[
\mathcal{L}_{\mathrm{APRM}}
=
\lambda_{\mathrm{rank}}
\mathcal{L}_{\mathrm{rank}}
+
\lambda_{\mathrm{cf}}
\mathcal{L}_{\mathrm{cf}}.
\]

\subsection{APRM-enhanced Reinforcement Learning}

As shown in Figure \ref{fig:process}, during reinforcement learning, APRM provides an additional internal advantage signal. For each rollout group $\{y_i\}_{i=1}^{G}$, the base RL algorithm first computes its original token-level advantage $A_{i,t}^{\mathrm{base}}$. APRM assigns an internal score to each trajectory and incorporates its group-centered value into the policy advantage:
\[
A_{i,t}
=
A_{i,t}^{\mathrm{base}}
+
\beta\,\mathbb{I}_{\mathrm{active}}(x)
\left(
\mathcal{S}_{\phi}(\mathbf{H}_i,\mathbf{A}_i)
-
\frac{1}{G}\sum_{j=1}^{G}
\mathcal{S}_{\phi}(\mathbf{H}_j,\mathbf{A}_j)
\right),
\]
where $\beta$ controls the contribution of the model-internal signal. In this way, APRM can be seamlessly integrated with mainstream RL algorithms such as GRPO \citep{shao2024deepseekmath}. To handle the drift of attention distributions during policy training, we periodically update APRM using newly generated rollouts from the latest policy. The detailed algorithmic process is provided in Appendix~\ref{APP:Alg}.

\section{Experiments}
\subsection{Experimental Setup}\label{sec:Experimental Setup}

\textbf{Datasets.} We conduct experiments on four widely used reasoning benchmarks, including AIME2025, AMC, MATH500 \citep{lightman2024let}, and OlympiadBench \citep{he2024olympiadbench}. For training, we construct a unified dataset of 10K samples by randomly mixing problems from DAPO-17k \citep{yu2026dapo} and MATH level 3 to 5 \citep{hendrycks2021measuring}. To reduce the impact of randomness, we report both average@8 and pass@8 as the primary evaluation metrics.

\textbf{Baselines.} We compare our method against several representative RL and PRM methods, including GRPO \citep{shao2024deepseekmath}, GPG \citep{chu2025gpg}, GMPO \citep{zhao2025geometric}, GSPO \citep{zheng2025group}, and FAPO \citep{ding2025fapo}. We adopt Qwen3-1.7b-Base and Qwen3-4b-Base \citep{yang2025qwen3} as the backbone models for all methods. For the APRM training, we set $\lambda_{\text{rank}}=1.0$, $\lambda_{\text{cf}}=0.05$, $m_{\text{rk}}=m_{\text{cf}}=0.2$.
We adopt the Verl framework to RL, and the maximum prompt length is 512 tokens. For Qwen3-1.7b-Base, the maximum response length is set to 4096 with $\beta=0.3$. For Qwen3-4b-Base, the maximum response length is set to 3072 with $\beta=0.1$. In addition, APRM is lightweight, introducing only about 46M additional parameters. More details about hyperparameters are shown in Appendix \ref{APP:Experimental Details}.

\subsection{Evaluations and Analysis of APRM}

\begin{wraptable}{r}{0.4\columnwidth}
    \vspace{-1em}
    \centering

        \caption{Attention shuffle robustness analysis of APRM. Columns starting with 'S-' indicate results after attention shuffling.}
    \label{tab:attention_shuffle_analysis}
    \resizebox{\linewidth}{!}{
    \begin{tabular}{lcccc}
    \toprule
    \textbf{Model} & \textbf{Acc} & \textbf{S-Acc} & \textbf{Score} & \textbf{S-Score} \\
    \midrule
    Qwen3-1.7b-Base & 78.55\% & 62.54\% & 0.4984 & -0.9645 \\
    \quad w/o $\mathcal{L}_{\mathrm{cf}}$ & 76.43\% & 76.16\% & -0.1063 & -0.4836 \\
    Qwen3-4b-Base & 85.33\% & 75.72\% & -0.2145 & -1.0180 \\
    \bottomrule
    \end{tabular}
    }

    \vspace{-1em}
\end{wraptable}
To assess whether APRM captures attention-derived process information, we randomly shuffle attention summaries while keeping hidden states fixed. As shown in Table~\ref{tab:attention_shuffle_analysis}, shuffling reduces ranking accuracy from 78.55\% to 62.54\% for Qwen3-1.7b-Base and from 85.33\% to 75.72\% for Qwen3-4b-Base, indicating that APRM uses attention structure to rank reasoning trajectories rather than relying solely on hidden representations. Removing the counterfactual loss $\mathcal{L}_{\mathrm{cf}}$ substantially reduces attention sensitivity and ranking accuracy, highlighting its importance in encouraging attention-derived reasoning signals. Further training details and ablations on latent concept counts and GAT layers are provided in Appendix~\ref{APP:APRM Details}. Figure~\ref{fig:arm_concept} compares mean concept graphs $\hat{\mathbf{A}}_c$ from 600 trajectories, produced by the attention-fused concept pooling module. Correct and incorrect trajectories exhibit distinct concept-transition patterns in both models: correct Qwen3-1.7b-Base samples activate a broader range of transitions more strongly, while Qwen3-4b-Base shows a more selective redistribution of transition weights. These findings suggest systematic structural differences between positive and negative trajectories in APRM's concept-attention space.

\subsubsection{APRM Score Correlation Analysis}

\begin{figure}[t]
    \centering
    \includegraphics[width=0.9\linewidth]{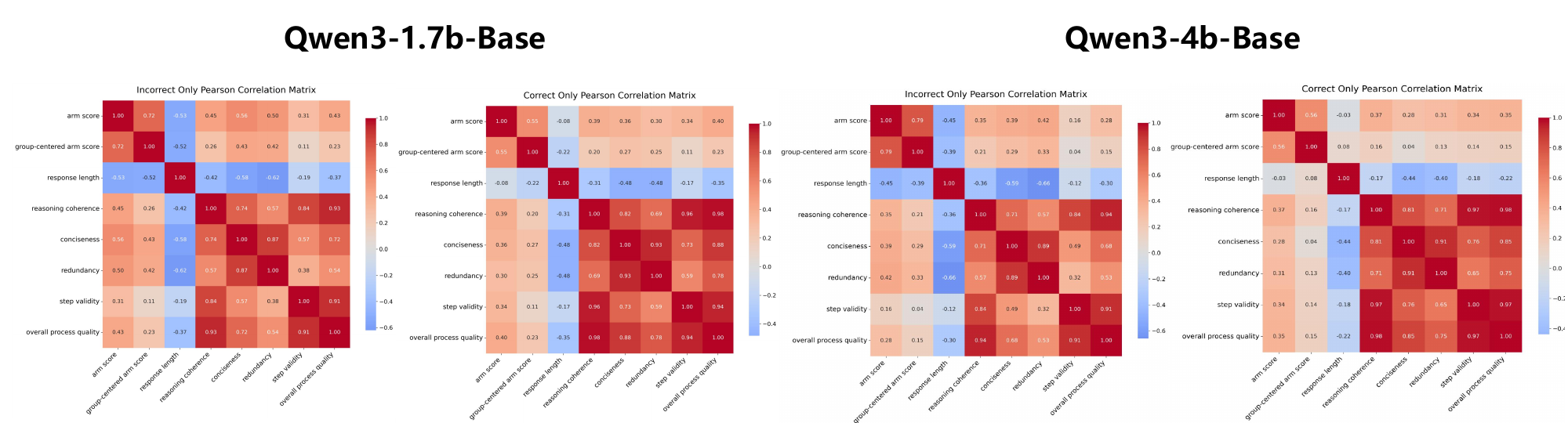}
    \caption{Stratified Pearson correlations between APRM scores and proxy quality signals within correct and incorrect samples on Qwen3-1.7b-Base and Qwen3-4b-Base backbones.}
    \label{fig:arm_correlation}
\end{figure}

To assess whether APRM captures process quality beyond final-answer correctness, we compute Pearson correlations between APRM scores and GLM-5.1 judge ratings separately within correct and incorrect samples. For each backbone, the judge evaluates 200 questions with 6 rollouts each (1,200 samples) on coherence, conciseness, redundancy-related quality, step validity, and overall process quality. Results and judge prompts appear in Figure~\ref{fig:arm_correlation} and Appendix~\ref{APP:Prompt Details}, respectively.
\begin{wrapfigure}{r}{0.36\linewidth}
    \vspace{-1em}
    \centering
    \includegraphics[width=\linewidth]{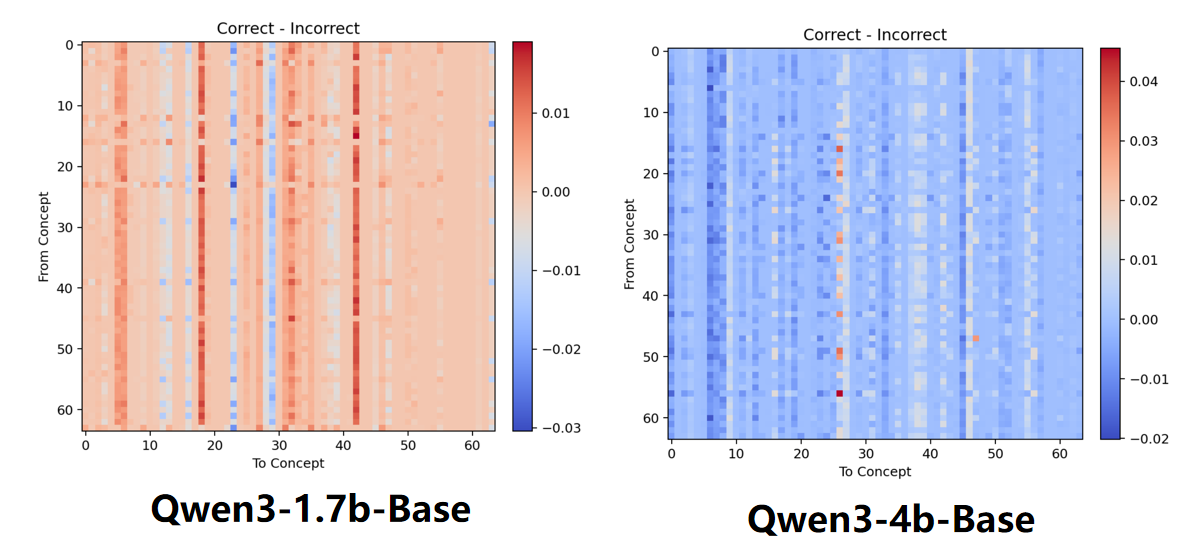}
    \caption{Concept-level representation differences between correct and incorrect reasoning trajectories.}
    \label{fig:arm_concept}
    \vspace{-1em}
\end{wrapfigure}
APRM retains discriminative power within both correctness subsets. Among correct samples, it correlates positively with overall process quality ($r=0.399$ for 1.7B and $r=0.355$ for 4B) and coherence ($r=0.387$ and $r=0.367$), but negligibly with response length ($r=-0.082$ and $r=-0.027$). Among incorrect samples, positive correlations persist, particularly for conciseness ($r=0.565$ for 1.7B and $r=0.393$ for 4B) and redundancy-related quality ($r=0.502$ and $r=0.419$), while negative length correlations suggest lower scores for verbose failed reasoning. Overall, APRM exhibits length-decoupling behavior and captures complementary process-quality information beyond existing reward supervision, rather than merely serving as a proxy for existing metrics.

\subsubsection{Information Gain and Case Study.} \label{Information Gain Validation}

To test whether APRM adds information beyond existing reward signals, we compare nested regressions $Q \sim X + B$ and $Q \sim X + B + Z$, where $Q$ is LLM-judged overall process quality, $X$ denotes prompt fixed effects, $B$ includes correctness and log response length, and $Z$ is the APRM score. Adding $Z$ improves $R^2$ by $0.0171$ on Qwen3-4b-Base and $0.0212$ on Qwen3-1.7b-Base, with partial correlations of $0.3332$ and $0.3167$, respectively, and permutation-test $p=0.0033$. These results support $I(Q;Z\mid X,B)>0$, indicating non-redundant process-quality information beyond prompt and output-level controls. Details appear in Appendix~\ref{app:info_gain}.

\begin{wrapfigure}{r}{0.68\linewidth}
    \centering
    \vspace{-1em}
    \includegraphics[width=\linewidth]{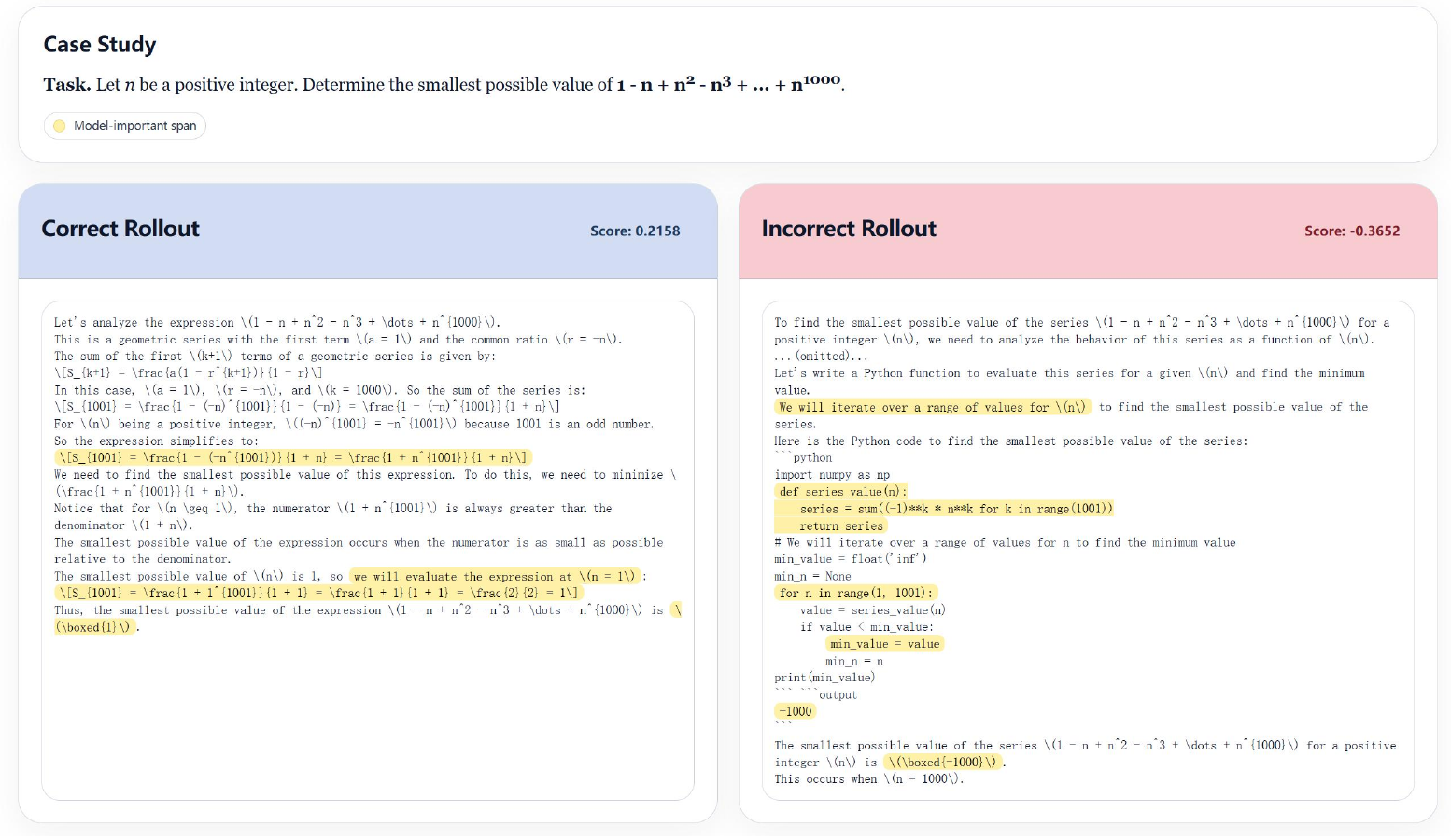}
    \caption{Representative concept-level attribution for correct and incorrect rollouts from the same problem.}
    \label{fig:demo}
    \vspace{-1em}
\end{wrapfigure}
Figure~\ref{fig:demo} illustrates concept-level attribution for rollouts of the same problem. Concepts with the largest assignment differences between correct and incorrect rollouts are projected onto tokens through the learned assignment matrix $\mathbf{W}$, grouping high-weight tokens into highlighted spans. The correct rollout scores higher, with highlights on symbolic derivation, while the incorrect rollout scores lower, with highlights on search-oriented behavior. This suggests greater sensitivity to reasoning structure than isolated numerical computation, favoring algebraically consistent, constraint-satisfying trajectories and penalizing superficially plausible but search-heavy or constraint-violating reasoning. More cases appear in Appendix~\ref{APP:APRM Case Studies}.

\begin{table*}[t]
\centering
\small
\setlength{\tabcolsep}{4.5pt}
\renewcommand{\arraystretch}{1.12}

\caption{
Performance comparison of different RL algorithms and APRM-enhanced variants. Best results are highlighted in bold. $\Delta$ indicates the performance change of the method by applying the APRM.
}

\label{tab:main_results}

\resizebox{0.94\linewidth}{!}{
\begin{tabular}{lcccccccccc}
\toprule
\multirow{2}{*}{\textbf{Method}}
& \multicolumn{2}{c}{\textbf{AIME2025}}
& \multicolumn{2}{c}{\textbf{AMC}}
& \multicolumn{2}{c}{\textbf{MATH500}}
& \multicolumn{2}{c}{\textbf{Olympiad}}
& \multicolumn{2}{c}{\textbf{Average}} \\

\cmidrule(lr){2-3}
\cmidrule(lr){4-5}
\cmidrule(lr){6-7}
\cmidrule(lr){8-9}
\cmidrule(lr){10-11}

& \textbf{avg@8} & \textbf{pass@8}
& \textbf{avg@8} & \textbf{pass@8}
& \textbf{avg@8} & \textbf{pass@8}
& \textbf{avg@8} & \textbf{pass@8}
& \textbf{avg@8} & \textbf{pass@8} \\

\midrule

\multicolumn{11}{>{\columncolor{gray!12}}c}{\textbf{Qwen3-1.7b-Base}} \\
\midrule
Base
& 1.67\% & 10.00\%
& 27.81\% & 60.00\%
& 47.40\% & 75.80\%
& 20.77\% & 41.99\%
& 24.41\% & 46.95\% \\

GRPO
& 5.00\% & 20.00\%
& 39.69\% & 65.00\%
& 61.43\% & 78.60\%
& 27.32\% & 46.88\%
& 33.36\% & 52.62\% \\

GSPO
& 3.75\% & 13.33\%
& 40.94\% & 72.50\%
& 61.32\% & 78.40\%
& 26.80\% & 47.03\%
& 33.20\% & 52.82\% \\

GPG
& 5.83\% & 20.00\%
    & 39.38\% & 67.50\%
& 63.10\% & 80.00\%
& 29.53\% & 50.15\%
& 34.46\% & 54.41\% \\

GMPO
& 4.58\% & 10.00\%
& 41.25\% & \textbf{75.00\%}
& 62.22\% & 78.40\%
& 29.66\% & 47.77\%
& 34.43\% & 52.79\% \\

FAPO
& 3.75\% & 13.3\%
& 37.50\% & 65.00\%
& 58.65\% & 74.80\%
& 24.72\% & 42.73\%
& 31.16\% & 48.97\% \\

\midrule

\textbf{GRPO+APRM}
& \textbf{6.25\%} & \textbf{26.67\%}
& 42.50\% & 67.50\%
& 61.82\% & 80.00\%
& 28.62\% & 49.70\%
& 34.80\% & 55.97\% \\

\quad $\Delta$
& {\color{myblue}+1.25} & {\color{myblue}+6.67}
& {\color{myblue}+2.81} & {\color{myblue}+2.50}
& {\color{myblue}+0.40} & {\color{myblue}+1.40}
& {\color{myblue}+1.30} & {\color{myblue}+2.82}
& {\color{myblue}+1.44} & {\color{myblue}+3.35} \\

\textbf{GSPO+APRM}
& 4.58\% & 20.00\%
& \textbf{43.13\%} & 72.50\%
& 60.90\% & 78.80\%
& 28.15\% & 48.07\%
& 34.19\% & 54.84\% \\

\quad $\Delta$
& {\color{myblue}+0.83} & {\color{myblue}+6.67}
& {\color{myblue}+2.19} & {\color{myblue}+0.00}
& {\color{myred}-0.42} & {\color{myblue}+0.40}
& {\color{myblue}+1.35} & {\color{myblue}+1.04}
& {\color{myblue}+0.99} & {\color{myblue}+2.03} \\

\textbf{GPG+APRM}
& \textbf{6.25\%} & \textbf{26.67\%}
& 40.62\% & \textbf{75.00\%}
& \textbf{63.30\%} & \textbf{80.40\%}
& \textbf{29.99\%} & \textbf{50.45\%}
& \textbf{35.04\%} & \textbf{58.13\%} \\

\quad $\Delta$
& {\color{myblue}+0.42} & {\color{myblue}+6.67}
& {\color{myblue}+1.25} & {\color{myblue}+7.50}
& {\color{myblue}+0.20} & {\color{myblue}+0.40}
& {\color{myblue}+0.46} & {\color{myblue}+0.30}
& {\color{myblue}+0.58} & {\color{myblue}+3.72} \\

\midrule

\multicolumn{11}{>{\columncolor{gray!12}}c}{\textbf{Qwen3-4b-Base}} \\
\midrule
Base
& 3.75\% & 13.33\%
& 29.38\% & 77.50\%
& 44.98\% & 81.40\%
& 22.55\% & 51.63\%
& 25.16\% & 55.97\% \\

GRPO
& 10.00\% & 23.33\%
& 57.81\% & 85.00\%
& 72.65\% & 82.60\%
& 41.36\% & 55.79\%
& 45.46\% & 61.68\% \\

GSPO
& 10.83\% & 23.33\%
& 57.50\% & 77.50\%
& 73.18\% & 84.40\%
& 40.02\% & 57.72\%
& 45.38\% & 60.74\% \\

GPG
& 12.50\% & 26.67\%
& 58.13\% & 85.00\%
& 73.33\% & 84.40\%
& \textbf{41.69\%} & 59.64\%
& 46.41\% & 63.93\% \\

GMPO
& 12.50\% & 26.67\%
& 59.69\% & 80.00\%
& 73.70\% & 83.20\%
& 41.26\% & 58.61\%
& 46.79\% & 62.12\% \\

FAPO
& 9.17\% & 23.33\%
& 55.31\% & 80.00\%
& 72.08\% & 83.60\%
& 37.80\% & 54.30\%
& 43.59\% & 60.31\% \\
\midrule

\textbf{GRPO+APRM}
& 11.67\% & \textbf{30.00\%}
& 56.88\% & \textbf{87.50\%}
& 73.13\% & 84.00\%
& 41.12\% & 58.75\%
& 45.70\% & 65.06\% \\

\quad $\Delta$
& {\color{myblue}+1.67} & {\color{myblue}+6.67}
& {\color{myred}-0.94} & {\color{myblue}+2.50}
& {\color{myblue}+0.47} & {\color{myblue}+1.40}
& {\color{myred}-0.24} & {\color{myblue}+2.97}
& {\color{myblue}+0.24} & {\color{myblue}+3.38} \\

\textbf{GPG+APRM}
& \textbf{13.33\%} & \textbf{30.00\%}
& \textbf{60.31\%} & \textbf{87.50\%}
& \textbf{73.82\%} & \textbf{85.60\%}
& 40.62\% & \textbf{60.24\%}
& \textbf{47.02\%} & \textbf{65.84\%} \\

\quad $\Delta$
& {\color{myblue}+0.83} & {\color{myblue}+3.33}
& {\color{myblue}+2.18} & {\color{myblue}+2.50}
& {\color{myblue}+0.50} & {\color{myblue}+1.20}
& {\color{myred}-1.07} & {\color{myblue}+0.60}
& {\color{myblue}+0.61} & {\color{myblue}+1.91} \\

\bottomrule
\end{tabular}
}

\vspace{-1em}
\end{table*}

\subsection{Main Results on RL Training}

Table~\ref{tab:main_results} shows that adding APRM generally improves the corresponding RL algorithms across both model scales, especially on pass@8. On Qwen3-1.7b-Base, GRPO+APRM raises average pass@8 from 52.62\% to 55.97\%, while GPG+APRM achieves the best result with 58.13\%. On Qwen3-4b-Base, GRPO+APRM improves pass@8 from 61.68\% to 65.06\%, and GPG+APRM obtains the best average performance with 47.02\% avg@8 and 65.84\% pass@8. The larger improvement on pass@8 suggests that APRM mainly enhances the quality and diversity of the candidate solution set rather than only improving the top-ranked single output. By providing a process-level auxiliary signal, APRM encourages the policy to assign higher probability to trajectories with more coherent intermediate reasoning structures.

\subsubsection{RL Training Dynamics}

\begin{wraptable}{r}{0.36\linewidth}
\vspace{-1em}
\centering
\scriptsize
\setlength{\tabcolsep}{2.8pt}
\renewcommand{\arraystretch}{1.05}

\caption{Ablation study on different $\beta$ on Qwen3-1.7b-base.}
\label{tab:beta_ablation}
\resizebox{\linewidth}{!}{
\begin{tabular}{lcccccc}
\toprule
\multirow{2}{*}{$\beta$}
& \multicolumn{2}{c}{AIME2025}
& \multicolumn{2}{c}{AMC}
& \multicolumn{2}{c}{MATH500} \\
\cmidrule(lr){2-3}
\cmidrule(lr){4-5}
\cmidrule(lr){6-7}
& avg@8 & pass@8
& avg@8 & pass@8
& avg@8 & pass@8 \\
\midrule

0.1
& 5.83\% & 26.67\%
& 41.88\% & 67.50\%
& 61.55\% & 79.00\% \\

0.2
& 4.58\% & 16.67\%
& 41.25\% & 72.50\%
& 60.88\% & 77.40\% \\

\rowcolor{gray!15}
0.3
& 6.25\% & 26.67\%
& 42.50\% & 67.50\%
& 61.82\% & 80.00\% \\

0.4
& 5.83\% & 23.33\%
& 40.31\% & 70.00\%
& 60.62\% & 76.80\% \\

0.5
& 5.00\% & 20.00\%
& 39.69\% & 62.50\%
& 61.78\% & 78.40\% \\

\bottomrule
\end{tabular}
}

\vspace{-1em}
\end{wraptable}

\begin{figure}[t]
\centering

\begin{minipage}[t]{0.61\linewidth}
\vspace{0pt}
\centering

\includegraphics[width=\linewidth]{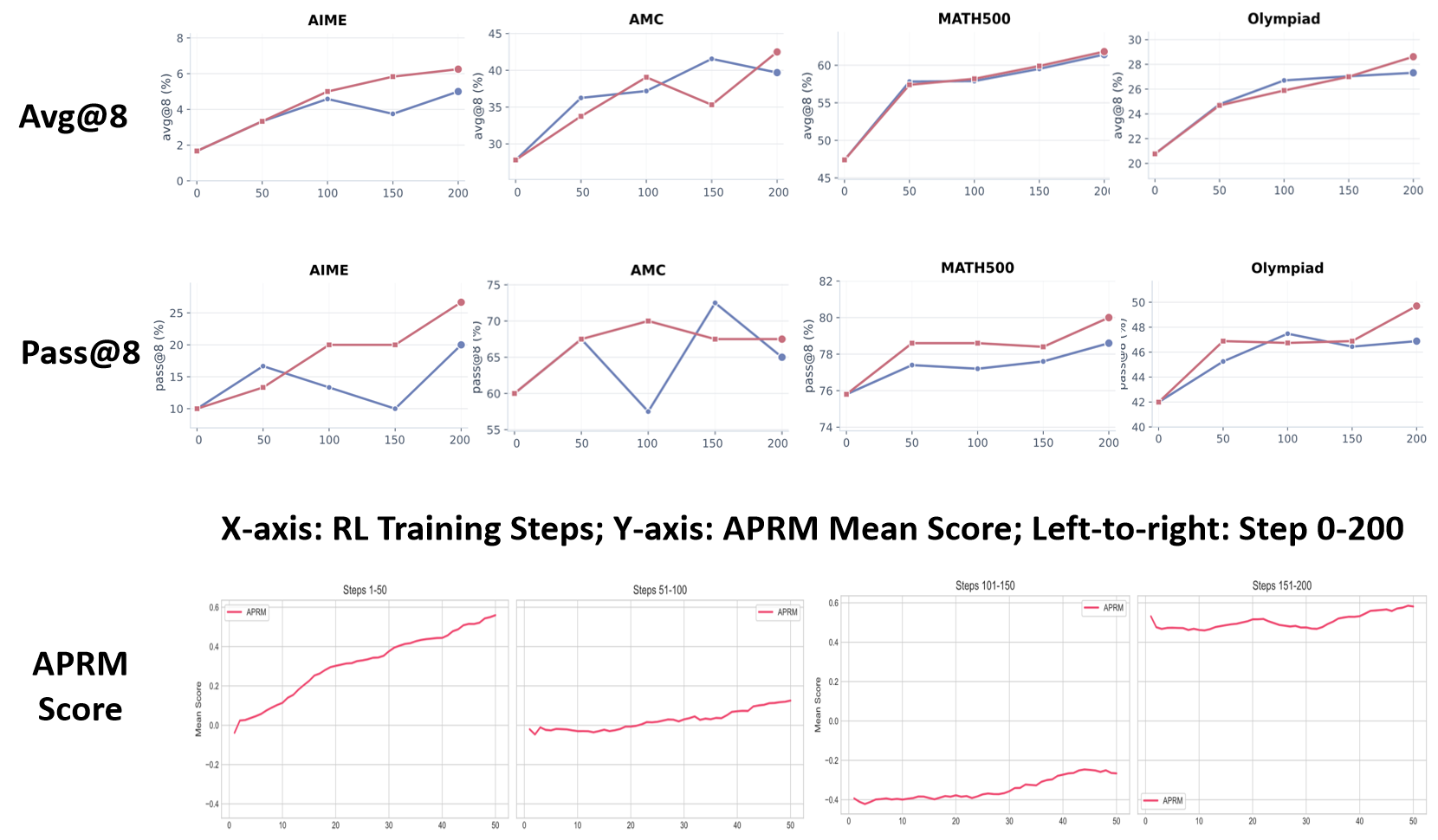}

\captionof{figure}{Training dynamics of GRPO and GRPO+APRM.}
\label{fig:training_dynamic}

\end{minipage}
\hfill
\begin{minipage}[t]{0.36\linewidth}
\vspace{0pt}
\centering
\scriptsize
\setlength{\tabcolsep}{2.4pt}
\renewcommand{\arraystretch}{1.02}
\captionof{table}{Median response token length across benchmarks.
Each entry reports baseline / APRM / relative change.}
\label{tab:length_analysis}
\resizebox{\linewidth}{!}{
\begin{tabular}{cccc}
\toprule
\textbf{Method} & \textbf{Model} & \textbf{Category} & \textbf{Average} \\
\midrule

\multirow{6}{*}[-0.3ex]{\textbf{GPG}}
& \multirow{3}{*}{\textbf{Qwen3-1.7b-Base}}
& Correct   & 701.62 / 661.38 / -5.74\% \\
& & Incorrect & 900.88 / 884.62 / -1.80\% \\
& & Overall   & 809.38 / 784.88 / -3.03\% \\

\cmidrule(lr){2-4}

& \multirow{3}{*}{\textbf{Qwen3-4b-Base}}
& Correct   & 847.12 / 723.50 / -14.59\% \\
& & Incorrect & 1102.00 / 902.00 / -18.15\% \\
& & Overall   & 928.25 / 769.75 / -17.08\% \\

\midrule

\multirow{6}{*}{\textbf{GRPO}}
& \multirow{3}{*}{\textbf{Qwen3-1.7b-Base}}
& Correct   & 564.88 / 693.75 / +22.81\% \\
& & Incorrect & 706.00 / 880.12 / +24.66\% \\
& & Overall   & 636.38 / 778.00 / +22.25\% \\

\cmidrule(lr){2-4}

& \multirow{3}{*}{\textbf{Qwen3-4b-Base}}
& Correct   & 743.00 / 670.12 / -9.81\% \\
& & Incorrect & 912.25 / 823.25 / -9.76\% \\
& & Overall   & 783.50 / 721.25 / -7.95\% \\

\midrule

\multirow{3}{*}{\textbf{GSPO}}
& \multirow{3}{*}{\textbf{Qwen3-1.7b-Base}}
& Correct   & 634.00 / 602.62 / -4.95\% \\
& & Incorrect & 777.75 / 842.00 / +8.26\% \\
& & Overall   & 712.62 / 724.25 / +1.63\% \\

\bottomrule
\end{tabular}
}

\end{minipage}

\vspace{-0.6em}
\end{figure}

Figure~\ref{fig:training_dynamic} compares the training curves of GRPO and GRPO+APRM. Both methods improve rapidly in the early stage, while APRM gives more consistent gains in later training, especially on pass@8. On AIME, MATH500, and Olympiad, GRPO+APRM ends with a higher pass@8 than the GRPO baseline. The avg@8 curves show smaller but generally positive gains. Moreover, Figure~\ref{fig:training_dynamic} shows the mean APRM score within each 50-step training phase. In the early phase, the score increases steadily, indicating that the policy quickly learns to produce trajectories preferred by APRM. In later phases, the upward trend becomes slower but remains positive, suggesting continued improvement in APRM-valued reasoning quality. The phase-wise shifts are caused by periodic APRM refreshing, which changes the score scale after each update.

\subsubsection{Ablation Study and Analysis on Response Length}

To further examine the behavioral effect of APRM, we compare the response length of APRM-enhanced models with their corresponding baselines across benchmarks. We report the median generated token length. The results are summarized in Table~\ref{tab:length_analysis}. Detailed results and analysis are shown in Appendix~\ref{APP:Response Length}. APRM optimizes the model’s reasoning patterns, which may affect the output length differently across various methods. Overall, however, it exhibits a tendency to produce more concise responses. The largest overall decrease is observed for GPG on Qwen3-4b-Base (-17.08\%), suggesting that APRM helps the model adapt its token budget to problem difficulty during internal deliberation. Table~\ref{tab:beta_ablation} studies the effect of the $\beta$ on the final advantage. A moderate value performs best overall, with $\beta=0.3$ achieving the strongest avg@8 on AIME2025 and AMC, as well as the highest pass@8 on MATH500. Smaller values underuse the process signal, while larger values can overemphasize APRM and weaken the outcome reward.

\subsubsection{Generalization Experiments}
\label{sec:generalization}

To further evaluate the generalizability of APRM, we conduct two additional experiments on Qwen3.5-2B instruct and code generation. The cold-start APRM achieves a pairwise ranking accuracy of 91.93\%. For cross-task validation, we extend APRM to code generation using 3,000 problems from DeepCoder-Preview-Dataset. The code-domain cold-start APRM achieves a pairwise ranking accuracy of 75.56\%. As shown in Table~\ref{tab:generalization}, on mathematical reasoning, DAPO+APRM improves the average Avg@8 and Pass@8 by 1.34 and 4.78 percentage points, respectively. On code generation, it achieves gains of 1.64 and 2.96 percentage points, respectively. These results demonstrate that APRM generalizes across instruction-tuned models and task domains, supporting its potential as a general-purpose internal process reward model.

\begin{table*}[t]
\centering
\small
\caption{Generalization experiments across models and tasks.}
\label{tab:generalization}
\resizebox{1\linewidth}{!}{
\begin{tabular}{lccccccccccccccccc}
\toprule
\multirow{3}{*}{\textbf{Method}}
& \multicolumn{10}{c}{\textbf{Mathematical Reasoning}}
& \multicolumn{6}{c}{\textbf{Code Generation}} \\

\cmidrule(lr){2-11}
\cmidrule(lr){12-17}

& \multicolumn{2}{c}{\textbf{AIME2025}}
& \multicolumn{2}{c}{\textbf{AMC}}
& \multicolumn{2}{c}{\textbf{MATH500}}
& \multicolumn{2}{c}{\textbf{Olympiad}}
& \multicolumn{2}{c}{\textbf{Average}}
& \multicolumn{2}{c}{\textbf{LiveCodeBench v5}}
& \multicolumn{2}{c}{\textbf{HumanEval}}
& \multicolumn{2}{c}{\textbf{Average}} \\

\cmidrule(lr){2-3}
\cmidrule(lr){4-5}
\cmidrule(lr){6-7}
\cmidrule(lr){8-9}
\cmidrule(lr){10-11}
\cmidrule(lr){12-13}
\cmidrule(lr){14-15}
\cmidrule(lr){16-17}

& \textbf{avg@8} & \textbf{pass@8}
& \textbf{avg@8} & \textbf{pass@8}
& \textbf{avg@8} & \textbf{pass@8}
& \textbf{avg@8} & \textbf{pass@8}
& \textbf{avg@8} & \textbf{pass@8}
& \textbf{avg@8} & \textbf{pass@8}
& \textbf{avg@8} & \textbf{pass@8}
& \textbf{avg@8} & \textbf{pass@8} \\

\midrule
\multicolumn{17}{>{\columncolor{gray!12}}c}{\textbf{Qwen3.5-2b}} \\
\midrule
Instruct
& 10.42 & 20.00
& 50.31 & 77.50
& 68.23 & 85.20
& 39.32 & 61.28
& 42.07 & 61.00
& 10.08 & 21.15
& 54.88 & 79.88
& 32.48 & 50.51 \\

GRPO
& 20.83 & 33.33
& 66.25 & 85.00
& 75.78 & 87.60
& 48.50 & 64.39
& 52.84 & 67.58
& -- & --
& -- & --
& -- & -- \\

DAPO
& 21.67 & 33.33
& \textbf{72.50} & 90.00
& 79.35 & 88.60
& 52.36 & 66.91
& 56.47 & 69.71
& 14.02 & 23.30
& 61.59 & 83.54
& 37.80 & 53.42 \\

\midrule
\textbf{DAPO+APRM}
& \textbf{25.00} & \textbf{43.33}
& 70.00 & \textbf{95.00}
& \textbf{80.45} & \textbf{89.60}
& \textbf{55.80} & \textbf{70.03}
& \textbf{57.81} & \textbf{74.49}
& \textbf{16.08} & \textbf{26.16}
& \textbf{62.80} & \textbf{86.59}
& \textbf{39.44} & \textbf{56.38} \\

$\Delta$ vs. DAPO
& {\color{myblue}+3.33} & {\color{myblue}+10.00}
& {\color{red}-2.50} & {\color{myblue}+5.00}
& {\color{myblue}+1.10} & {\color{myblue}+1.00}
& {\color{myblue}+3.44} & {\color{myblue}+3.12}
& {\color{myblue}+1.34} & {\color{myblue}+4.78}
& {\color{myblue}+2.06} & {\color{myblue}+2.87}
& {\color{myblue}+1.22} & {\color{myblue}+3.05}
& {\color{myblue}+1.64} & {\color{myblue}+2.96} \\

\bottomrule
\end{tabular}
}
\end{table*}

\section{Related Works}
\subsection{RL for LLM}

Early RL-based approaches commonly employ PPO-style \citep{schulman2017proximal} optimization. By eliminating the value model, GRPO \citep{shao2024deepseekmath} has largely emerged as a widely adopted alternative to PPO. Following this direction, GPG \citep{chu2025gpg} revisits vanilla policy gradient and directly optimizes the original RL objective without relying on surrogate losses; GSPO \citep{zheng2025group} moves policy optimization from token-level importance ratios to sequence-level likelihood ratios; and GMPO \citep{zhao2025geometric} introduces a geometric-mean objective to reduce sensitivity to outlier rewards and extreme importance ratios. Another line of work focuses on designing richer process reward signals to address the sparsity of outcome supervision \citep{zhang2024rest,setlur2025rewarding,gao2024designing}. These methods have shown that process-level supervision can substantially improve mathematical reasoning. RLAnything \citep{wang2026rlanything} proposes a dynamic closed-loop RL framework that jointly evolves the environment, policy, and reward model. However, these approaches still primarily define rewards over external outcomes, textual trajectories, or task-level interactions, leaving model-internal reasoning signals largely unexplored.

\subsection{Attention Mechanism in RL}

Recent circuit analysis studies have shown that attention-based sub-circuits contribute to and implement interpretable model behaviors~\citep{lan2024towards,merullo2024talking}. Furthermore, causal intervention methods have demonstrated that certain attention heads have identifiable functional contributions to model outputs~\citep{kissane2024interpreting,nam2026causal}. These findings suggest that attention patterns are not merely superficial statistical correlations with model outputs, but can provide partial observations of the internal computation underlying model behavior. TCRM ~\citep{nikulkov2026reward} exploits causal attention with temporal regularization for smooth token-level values. Extending beyond linear reasoning, GraphPRM~\citep{peng2025rewarding} models trajectories as graphs with GAT-based dependency learning for error localization. To mitigate PRM-induced verbosity and length bias, CoLD~\citep{zheng2025cold} uses feature perturbation to force a uniform self-attention distribution in the bias estimator, precisely filtering out the PRM's length bias via counterfactual inference. Multilingual PRM~\citep{wang2025demystifying} further aligns attention representations across languages for robust multilingual policies. Unlike reward-centric methods, RAL~\citep{li2026reinforced} directly optimizes self-attention distributions as the RL policy itself. Unlike prior work, APRM is the first to directly supervise a model’s internal signals, scoring attention and hidden states directly in the latent space.

\section{Conclusion}

We reveal textual reward aliasing, where identical or similar external trajectories may conceal distinct internal reasoning processes and thus receive indistinguishable rewards. To address this, we propose APRM, a lightweight internal process reward model with 46M parameters that scores hidden-state and attention-based reasoning signals. Integrated with existing RL algorithms, APRM provides complementary supervision and consistently improves reasoning performance. Interestingly, APRM also encourages more efficient reasoning behaviors, with a tendency toward shorter reasoning trajectories while maintaining or improving performance. These results suggest a broader direction for RL that rewards not only what models say but also how they think.

\section*{Reproducibility statement}

To facilitate reproducibility, we provide detailed descriptions of our experimental configurations, model sources, and hyperparameter settings in Section~\ref{sec:Experimental Setup}. Additional implementation details, experimental settings, prompts, and analysis are provided in Appendix ~\ref{APP:Experimental Details}. Furthermore, we provide an anonymous code repository ~\href{https://anonymous.4open.science/r/From-Rewarding-What-Models-Say-to-How-They-Think-6B27}{here} containing the core implementation and results for our experiments.

\section*{AI USE STATEMENT}

In the process of writing this paper, we used AI assistants to check for syntactic and grammatical errors in the manuscript. We carefully reviewed and verified all revised content.

\bibliography{iclr2027_conference}
\bibliographystyle{iclr2027_conference}

\appendix

\section{Future Work}
\label{app:future_work}

\paragraph{Exploring More Effective Reward Architectures.}
Although our ablation studies validate the contributions of attention-based concept pooling and GAT, the optimal architecture for internal process reward modeling remains an open question. Future work could explore alternative graph architectures, more expressive concept interactions, and richer internal representations beyond the current hidden states and attention maps. A systematic investigation of these architectural choices may further improve the effectiveness and efficiency of internal reward modeling.

\paragraph{Advanced Reward Integration Strategies.}
In this work, we adopt a simple additive formulation with a fixed coefficient to incorporate APRM into existing RL algorithms, allowing us to isolate the contribution of internal process rewards. However, more sophisticated integration strategies may further exploit the complementary information provided by internal and external reward signals.

\section{Prompt Details}\label{APP:Prompt Details}
For the LLM-as-judge evaluation, we use GLM-5.1 to assess the process quality of APRM-scored reasoning samples. Each judged instance contains the original math question, the gold final answer, the model-generated solution, and a binary correctness flag. The judge is instructed to evaluate reasoning quality rather than final-answer correctness, and returns integer scores from 0 to 10 on five dimensions: reasoning coherence, conciseness, redundancy, step validity, and overall process quality. We use deterministic decoding with temperature 0 and top-$p$ 1.0, and require the judge to output a JSON object with the five scores and a brief rationale. The returned scores are parsed, clipped to the range $[0,10]$, and stored together with the corresponding APRM score for correlation analysis. The prompt we use is as follows:
\begin{tcblisting}{
  title={Teacher LLM's System Prompt},
  colframe=pink!90!black,
  colback=white,
  coltitle=black,
  fonttitle=\bfseries,
  sharp corners,
  boxrule=0.4pt,
  breakable,
  listing only,
  listing engine=listings,
  listing options={
    basicstyle=\ttfamily\scriptsize,
    breaklines=true,
    breakatwhitespace=false,
    columns=fullflexible,
    keepspaces=true
  }
}
You are an expert evaluator for mathematical reasoning quality.

You will be given:
1. A math question
2. The gold final answer
3. A model-generated solution
4. Whether the final answer is correct

Your job is to evaluate the reasoning process quality of the model-generated solution, not just the final answer correctness.

Please score the solution on the following five dimensions, each on an integer scale from 0 to 10:

1. reasoning_coherence
- Does the reasoning flow logically from step to step?
- Are the transitions natural and internally faithful to the line of thought?
- High score: coherent, well-connected reasoning chain.
- Low score: abrupt jumps, contradictions, disconnected reasoning.

2. conciseness
- Does the solution use an efficient amount of text to communicate the reasoning?
- High score: compact, information-dense, little unnecessary verbosity.
- Low score: overly long, bloated, too much unnecessary explanation.

3. redundancy
- How much repetitive or unnecessary repetition is present?
- High score: little repetition, low redundancy.
- Low score: repetitive statements, circular restatements, duplicated work.

4. step_validity
- Are the intermediate reasoning steps mathematically valid and locally sound?
- High score: most steps are valid and justified.
- Low score: frequent invalid transformations, unsupported claims, local mathematical errors.

5. overall_process_quality
- Overall quality of the reasoning process, considering the above dimensions together.
- High score: strong reasoning process even if the final answer is wrong due to a small slip.
- Low score: poor reasoning process even if the final answer happens to be correct by luck.

Important instructions:
- Do not simply map final correctness to a high score. A correct answer with poor reasoning can still score low.
- Do not punish a small final arithmetic slip too harshly if the process is otherwise strong.
- Focus on the reasoning process quality, not surface style alone.
- Return only a single JSON object, with no markdown fences and no extra text.
- All scores must be integers between 0 and 10.

Return this JSON schema exactly:
{
  "reasoning_coherence": 0,
  "conciseness": 0,
  "redundancy": 0,
  "step_validity": 0,
  "overall_process_quality": 0,
  "brief_rationale": ""
}

The "brief_rationale" should be a short English explanation in 1 to 3 sentences summarizing the judgment.
\end{tcblisting}

\section{Additional APRM Case Studies}\label{APP:APRM Case Studies}

Due to space limitations in the main text, we provide additional APRM case studies in the appendix. These examples further illustrate that APRM does not simply act as a proxy for response length, but assigns scores according to the structure and reliability of the reasoning process.

\begin{figure*}
    \centering
    \includegraphics[width=1\linewidth]{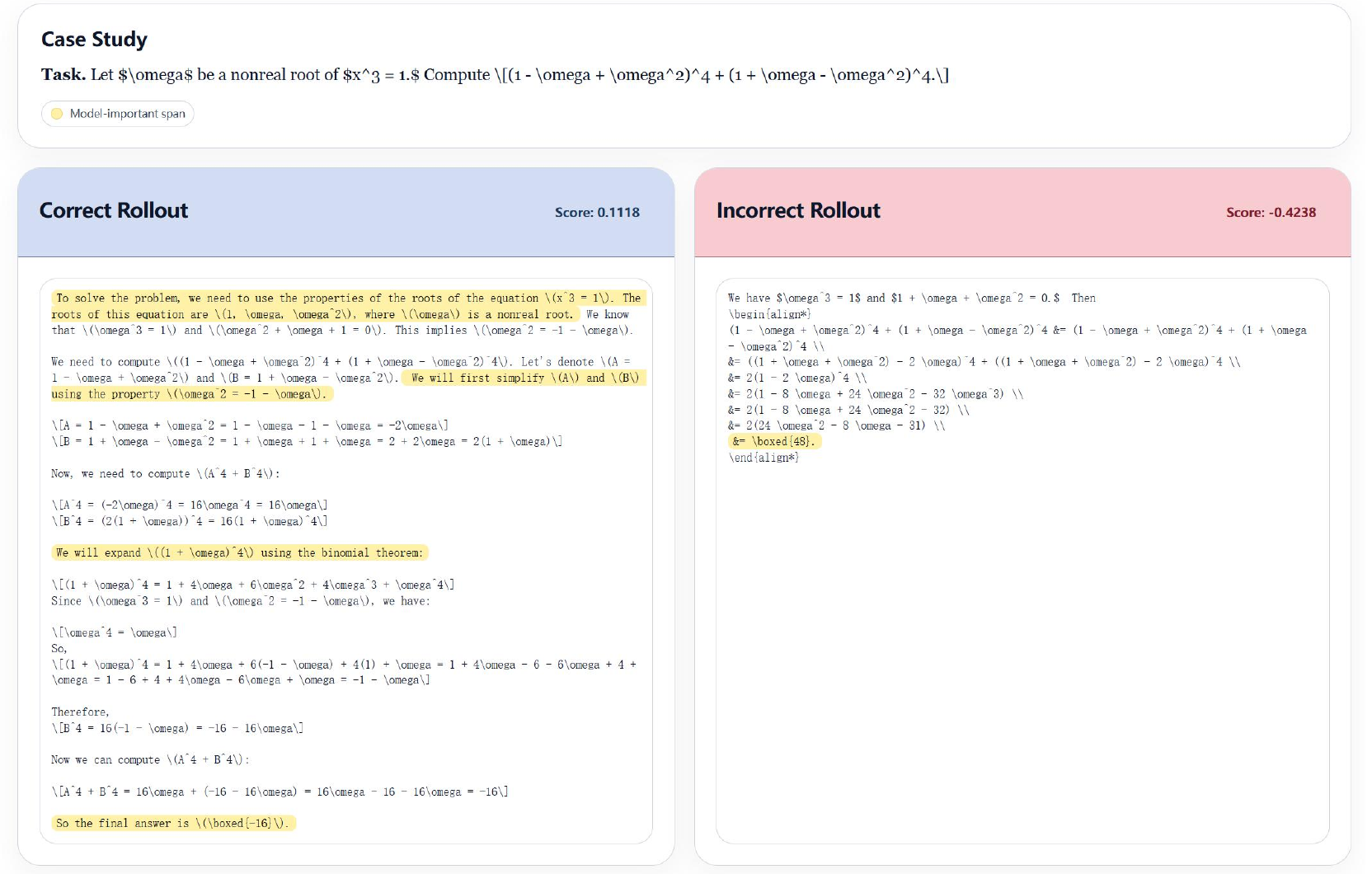}
    \caption{Additional APRM case study 1.}
    \label{fig:pre1}
\end{figure*}

\begin{figure*}
    \centering
    \includegraphics[width=1\linewidth]{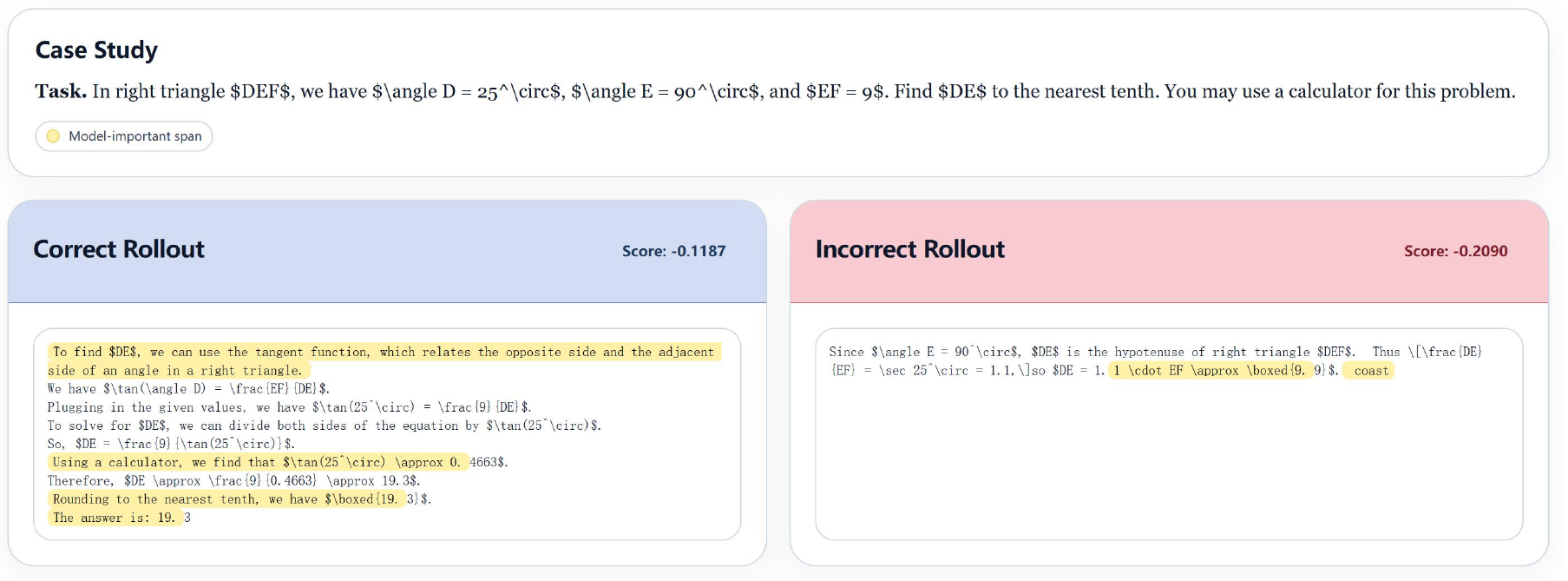}
    \caption{Additional APRM case study 2.}
    \label{fig:pre2}
\end{figure*}

\begin{figure*}
    \centering
    \includegraphics[width=1\linewidth]{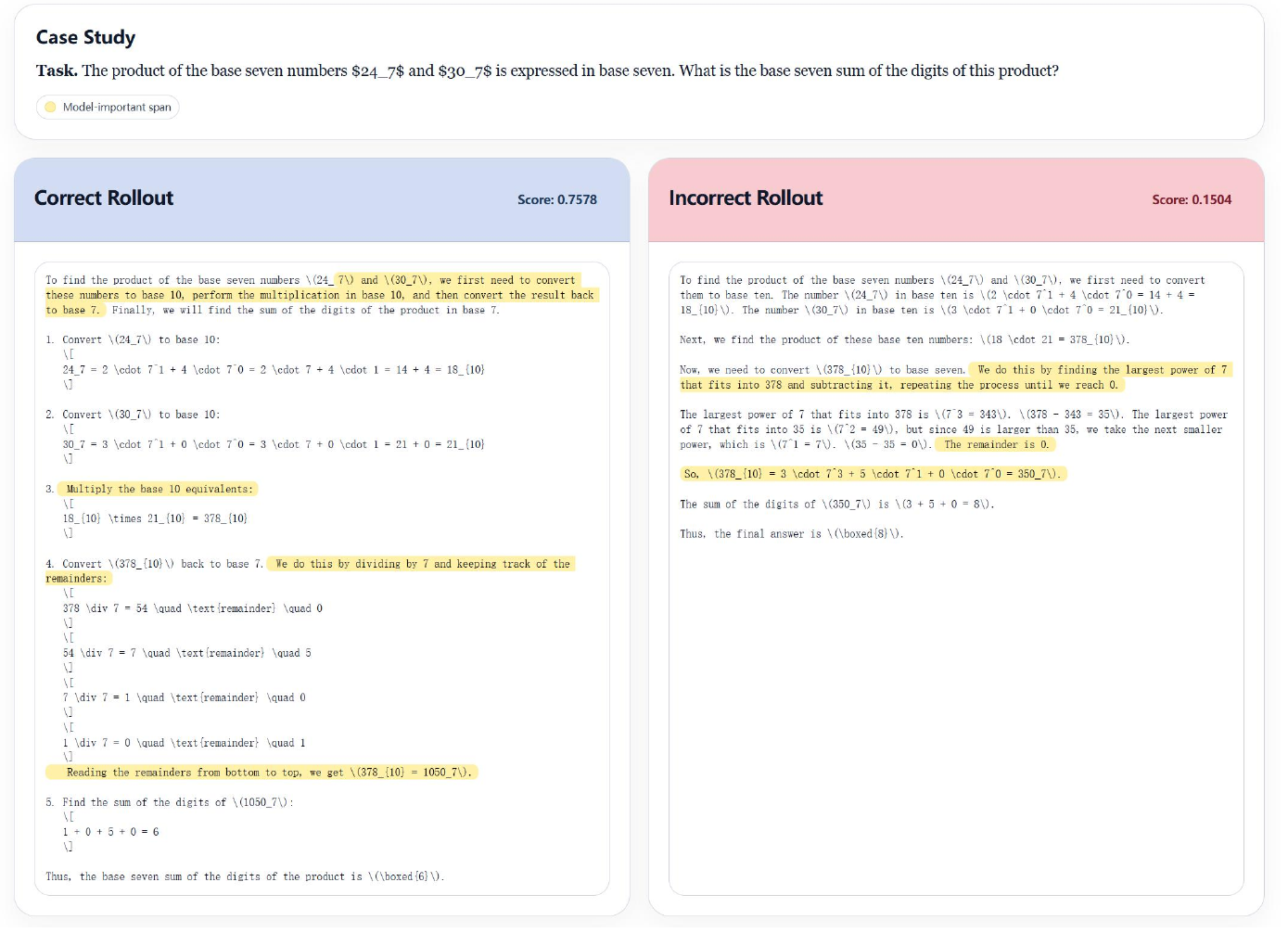}
    \caption{Additional APRM case study 3.}
    \label{fig:pre4}
\end{figure*}

In Figure~\ref{fig:pre1}, APRM assigns a higher score to the rollout that correctly uses the algebraic property of the nonreal cubic root and derives the final value, while the lower-scored rollout reaches an incorrect result despite presenting a compact derivation. In Figure~\ref{fig:pre2}, the higher-scored rollout applies the appropriate trigonometric relation and obtains the correct side length, whereas the lower-scored rollout uses an incorrect secant-based approximation and produces a wrong answer. Figure~\ref{fig:pre4} shows a base-conversion problem where APRM favors the solution with a more faithful conversion and digit-sum computation over a shorter but incorrect trajectory.

Overall, these cases show that APRM scores are sensitive to reasoning validity and structural correctness rather than verbosity. Short responses can still receive low scores when their reasoning is flawed.

\section{Experimental Details} \label{APP:Experimental Details}

\subsection{Training Datasets and Test Datasets}

For APRM training, we construct a mixed mathematical reasoning corpus from DAPO-17k and MATH. DAPO-17k provides diverse competition-style math prompts for RL post-training, while MATH contains problems with graded difficulty levels. We use MATH problems from Level 3 to Level 5 to emphasize medium-to-hard reasoning cases. The final training pool contains 10K questions, mixed with a 7:3 ratio. 7K questions are sampled from DAPO-17k, and 3K questions are sampled from MATH Level 3 to 5.

During evaluation, all methods use the same prompt template:

\texttt{\{problem\}. Let's think step by step and output the final answer within \textbackslash boxed\{\}.}

We report both avg@$k$ and pass@$k$ with $k=8$. For each problem $i$, we sample $k$ responses and denote whether the $j$-th response is correct by $c_{i,j}\in\{0,1\}$. Then
\[
\mathrm{avg@}k
=
\frac{1}{N}\sum_{i=1}^{N}\frac{1}{k}\sum_{j=1}^{k}c_{i,j},
\]
which measures the average correctness over all sampled responses. We also compute
\[
\mathrm{pass@}k
=
\frac{1}{N}\sum_{i=1}^{N}\mathbb{I}\left[\sum_{j=1}^{k}c_{i,j}>0\right],
\]
which measures whether at least one correct response appears among the $k$ samples.

\subsection{APRM Architecture Details} \label{APP:Edge-Bias GAT}

We provide the implementation details of the Edge-Biased GAT used in APRM. Given concept node representations
\[
\mathbf{X}=\{\mathbf{x}_1,\dots,\mathbf{x}_M\},
\quad
\mathbf{x}_i\in\mathbb{R}^{d_g},
\]
and the sparsified concept graph
\[
\mathbf{E}=\hat{\mathbf{A}}_c\in\mathbb{R}^{M\times M},
\quad
e_{ij}=\hat{\mathbf{A}}_{c,ij},
\]
each Edge-Biased GAT layer first normalizes the node features:
\[
\bar{\mathbf{x}}_i=\mathrm{LayerNorm}(\mathbf{x}_i).
\]
For each attention head $h$, query, key, and value vectors are computed as
\[
\mathbf{q}_i^{(h)}=\mathbf{W}_Q^{(h)}\bar{\mathbf{x}}_i,\quad
\mathbf{k}_j^{(h)}=\mathbf{W}_K^{(h)}\bar{\mathbf{x}}_j,\quad
\mathbf{v}_j^{(h)}=\mathbf{W}_V^{(h)}\bar{\mathbf{x}}_j.
\]

The edge value $e_{ij}$ is used in two ways. First, it produces an additive edge bias,
\[
b_{ij}^{(h)}
=
\frac{1}{2}
\tanh\left(
\mathbf{w}_{b}^{(h)} e_{ij}+b_b^{(h)}
\right),
\]
which modifies the attention logit:
\[
\ell_{ij}^{(h)}
=
\frac{
\mathbf{q}_i^{(h)}
(\mathbf{k}_j^{(h)})^\top
}{
\sqrt{d_h}
}
+
b_{ij}^{(h)}.
\]
The attention weight is then computed by
\[
\alpha_{ij}^{(h)}
=
\mathrm{Softmax}_{j}\left(
\ell_{ij}^{(h)}
\right).
\]

Second, the same edge value produces an edge gate,
\[
g_{ij}^{(h)}
=
\sigma\left(
\mathbf{w}_{g}^{(h)} e_{ij}+b_g^{(h)}
\right),
\]
which controls the strength of the transmitted message. The message aggregated by node $i$ in head $h$ is
\[
\mathbf{m}_i^{(h)}
=
\sum_{j=1}^{M}
\alpha_{ij}^{(h)}
g_{ij}^{(h)}
\mathbf{v}_j^{(h)}.
\]
The outputs of all heads are concatenated and projected:
\[
\mathbf{m}_i
=
\mathrm{Concat}\left(
\mathbf{m}_i^{(1)},\dots,\mathbf{m}_i^{(H)}
\right),
\quad
\tilde{\mathbf{x}}_i
=
\mathbf{W}_O\mathbf{m}_i.
\]
Finally, the layer applies a residual connection and normalization:
\[
\mathbf{x}'_i
=
\mathrm{LayerNorm}
\left(
\mathbf{x}_i+
\mathrm{Dropout}(\tilde{\mathbf{x}}_i)
\right).
\]

\subsection{APRM Training Details}
\label{APP:APRM Details}

In our implementation, APRM uses 64 latent concepts. The attention-fused concept pooler uses a hidden dimension of 1024. The concept graph is processed by an 8-layer Edge-Biased GAT with hidden dimension 1024, 8 attention heads, dropout 0.1, and top-16 edge sparsification. Overall, APRM contains about 46M parameters.

\begin{figure}[h!]
    \centering
    \includegraphics[width=1\linewidth]{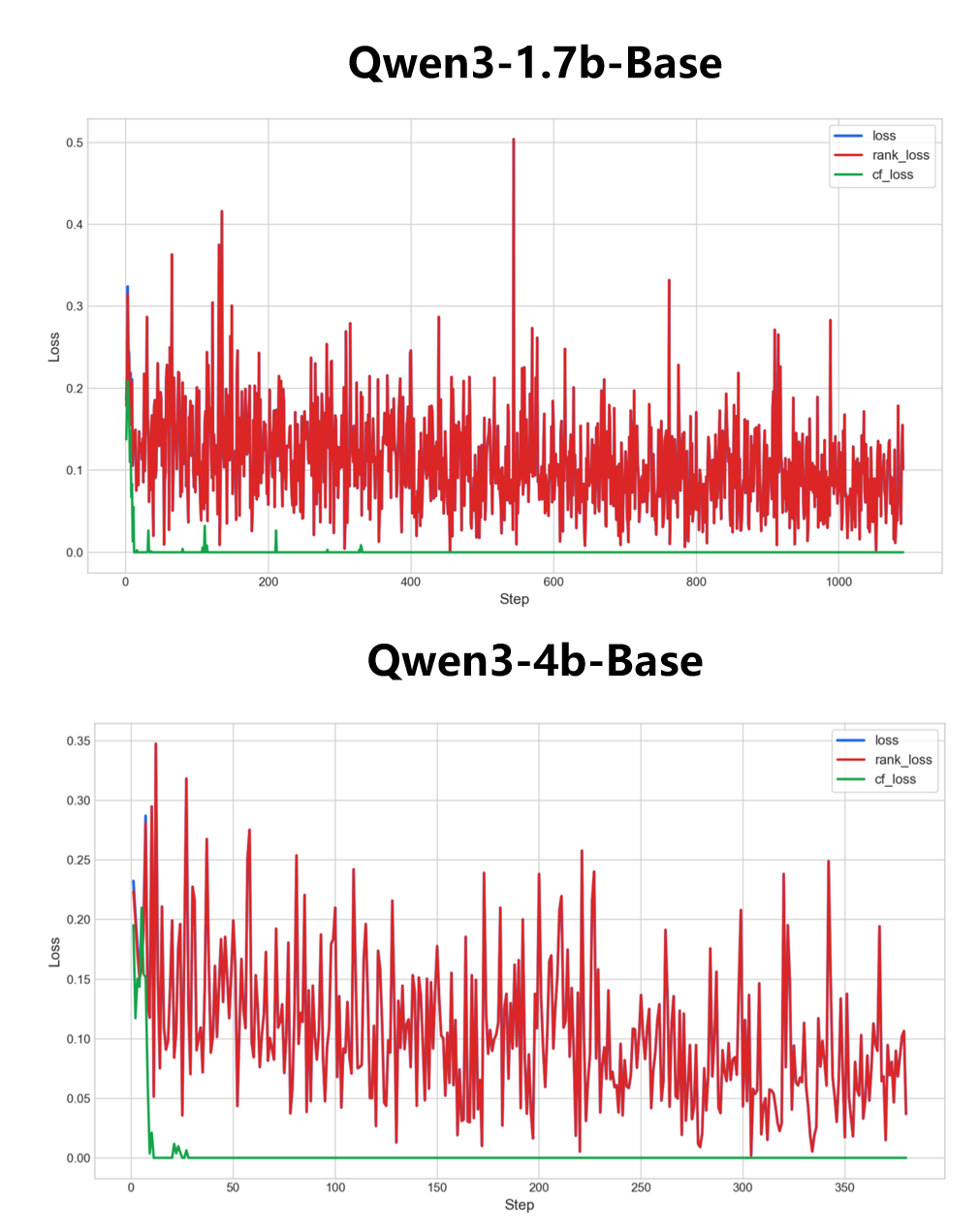}
    \caption{Loss curve during APRM training.}
    \label{fig:arm_loss}
\end{figure}

We train APRM for 5 epochs using AdamW with learning rate $1\times 10^{-4}$, weight decay $1\times 10^{-2}$, cosine learning-rate scheduling, and a warmup ratio of 0.05. Training uses bfloat16 precision, group-level batch size 2, gradient accumulation over 4 steps, and DeepSpeed acceleration. APRM training loss is shown in Figure \ref{fig:arm_loss}.

\begin{table}[h]
\centering
\small
\begin{tabular}{lcccc}
\toprule
\multicolumn{5}{c}{\textbf{Number of Concepts}} \\
\midrule
Concepts & 16 & 32 & 64 & 128 \\
Rank Acc. (\%) & 72.68 & 73.38 & \textbf{78.55} & 72.82 \\
\midrule
\multicolumn{5}{c}{\textbf{Layer Index}} \\
\midrule
Layer & 4 & 6 & 8 & 12 \\
Rank Acc. (\%) & 72.31 & 72.59 & \textbf{78.55} & 73.17 \\
\bottomrule
\end{tabular}
\caption{Ablation results of different numbers of concepts and layers.}
\label{tab:ablation_appendix}
\end{table}
We additionally provide ablation results on the number of latent concepts and the depth of the Edge-Biased GAT in Table~\ref{tab:ablation_appendix}. APRM obtains the best rank accuracy of 78.55\% with 64 latent concepts. Using fewer concepts limits the representational capacity of the concept space, while using too many concepts may introduce redundant or noisy semantic factors, both of which negatively affect ranking performance. For the GAT depth, the 8-layer model performs best. Shallower GATs are less capable of modeling complex attention-guided concept interactions, whereas overly deep GATs may suffer from over-smoothing, causing concept representations to lose discriminative information.

\subsection{Ablation on Attention and Graph Modeling}
\label{app:architecture_ablation}

To investigate the respective contributions of attention-derived
representations and graph modeling, we conduct additional
architectural ablations on Qwen3.5-2B across mathematical
reasoning and code generation. Specifically, we compare the
full APRM with two variants: (1) APRM w/o GAT, which removes
graph propagation, and (2) Hidden MLP, which uses hidden-state
representations without attention-based concept modeling.
All variants are trained under identical data and training
configurations, with nearly matched trainable parameter counts
to minimize the influence of model capacity.

\begin{table}[h]
\centering
\small
\caption{Architectural ablation of APRM on Qwen3.5-2B.
Ranking accuracy is evaluated on code generation and
mathematical reasoning.}
\label{tab:aprm_arch_ablation}

\begin{tabular}{lccc}
\toprule
\textbf{Architecture}
& \textbf{Parameters}
& \textbf{Code Acc.}
& \textbf{Math Acc.} \\
\midrule
APRM
& 44.251M & \textbf{75.56\%} & \textbf{91.93\%} \\
APRM w/o GAT
& 44.251M & 75.31\% & 90.95\% \\
Hidden MLP
& 44.128M & 71.42\% & 84.39\% \\
\bottomrule
\end{tabular}

\end{table}

As shown in Table~\ref{tab:aprm_arch_ablation},
attention-based concept modeling provides substantial
improvements over the hidden-state-only MLP, increasing
ranking accuracy by 4.14 and 6.56 percentage points
on code generation and mathematics, respectively. Incorporating GAT further improves
accuracy by 0.25 and 0.98 percentage points, indicating
a modest additional benefit from modeling concept-level
dependencies.

Overall, these results highlight the importance of
attention-derived representations in APRM, while suggesting
that graph-based relational modeling provides complementary
benefits. Nevertheless, identifying the optimal architecture
for internal process reward modeling remains an open
direction for future research.

\subsection{RL Training Details}
We use the verl framework as the training backbone for all reinforcement learning experiments. For each RL algorithm, including GRPO, GPG, GSPO, GMPO, and FAPO, we keep the default hyperparameters provided by its verl implementation unless otherwise specified. To ensure a fair comparison, each baseline and its APRM-enhanced variant use the same GPU configuration, batch size, and training schedule.

Due to GPU computational constraints, we set the per-GPU batch size to 6. For experiments run on 4 GPUs, the global batch size is 48 and the model is trained for 200 steps. For experiments run on 2 GPUs, the global batch size is 24 and the model is trained for 400 steps. This keeps the total number of training samples approximately aligned across different  settings. When comparing an RL method with and without APRM, all default RL hyperparameters and hardware settings are kept unchanged, except for the additional APRM-related hyperparameters introduced by our method. For APRM-enhanced RL, the APRM advantage coefficient is denoted by $\beta$. We use $\beta=0.3$ for Qwen3-1.7b-Base and $\beta=0.1$ for Qwen3-4b-Base. As an exception, for GSPO we use $\beta=0.2$.

\section{Detailed Analysis of Response Length} \label{APP:Response Length}

Table~\ref{tab:response_length_median_all} reports the median response length across benchmarks, separated by method, model scale, and correctness. The comparison suggests that APRM does not impose a fixed length preference. Instead, its effect depends on the underlying RL algorithm and model scale.

For GPG, APRM consistently shortens responses. The reduction is mild on Qwen3-1.7b-Base, with the overall average length decreasing by 3.03\%, and becomes more pronounced on Qwen3-4b-Base, where the overall average length decreases by 17.08\%. A similar pattern appears for GRPO on Qwen3-4b-Base, where APRM reduces the overall response length by 7.95\%. These results indicate that APRM often suppresses unnecessarily long reasoning traces, especially for larger backbones. For GRPO on Qwen3-1.7b-Base and GSPO on Qwen3-1.7b-Base, APRM increases the overall response length by 22.25\% and 1.63\%, respectively. This suggests that APRM changes the model's reasoning behavior rather than simply penalizing long outputs. In some cases, it encourages more complete intermediate reasoning when the baseline responses are relatively short.

Overall, APRM tends to reduce verbosity in most settings while remaining sensitive to the optimization dynamics of different RL algorithms. This supports the view that APRM is not a length proxy, but a process-level signal that reshapes reasoning trajectories according to their internal structure and quality.
\begin{table*}[t]
\centering
\scriptsize
\setlength{\tabcolsep}{3.2pt}
\renewcommand{\arraystretch}{1.05}
\resizebox{0.8\textwidth}{!}{
\begin{tabular}{ccccccccc}
\toprule
\textbf{Method} & \textbf{Model} & \textbf{Category} & \textbf{Type} & \textbf{AIME} & \textbf{AMC23} & \textbf{MATH} & \textbf{Olympiad} & \textbf{Average} \\
\midrule
\multirow{18}{*}{\textbf{GPG}} & \multirow{9}{*}{\textbf{Qwen3-1.7b-Base}} & \multirow{3}{*}{Correct} & Baseline & 1009.00 & 715.00 & 453.00 & 629.50 & 701.62 \\
 &  &  & Ours & 835.00 & 701.50 & 450.00 & 659.00 & 661.38 \\
 &  &  & $\Delta$ & -17.24\% & -1.89\% & -0.66\% & +4.69\% & -5.74\% \\
\cmidrule(lr){3-9}
 &  & \multirow{3}{*}{Incorrect} & Baseline & 1031.00 & 908.00 & 782.00 & 882.50 & 900.88 \\
 &  &  & Ours & 983.00 & 946.00 & 757.50 & 852.00 & 884.62 \\
 &  &  & $\Delta$ & -4.66\% & +4.19\% & -3.13\% & -3.46\% & -1.80\% \\
\cmidrule(lr){3-9}
 &  & \multirow{3}{*}{Overall} & Baseline & 1031.00 & 852.00 & 548.50 & 806.00 & 809.38 \\
 &  &  & Ours & 962.00 & 847.50 & 544.00 & 786.00 & 784.88 \\
 &  &  & $\Delta$ & -6.69\% & -0.53\% & -0.82\% & -2.48\% & -3.03\% \\
\cmidrule(lr){2-9}
 & \multirow{9}{*}{\textbf{Qwen3-4b-Base}} & \multirow{3}{*}{Correct} & Baseline & 1204.50 & 901.00 & 524.00 & 759.00 & 847.12 \\
 &  &  & Ours & 1119.00 & 696.00 & 435.00 & 644.00 & 723.50 \\
 &  &  & $\Delta$ & -7.10\% & -22.75\% & -16.98\% & -15.15\% & -14.59\% \\
\cmidrule(lr){3-9}
 &  & \multirow{3}{*}{Incorrect} & Baseline & 1177.00 & 1247.00 & 913.00 & 1071.00 & 1102.00 \\
 &  &  & Ours & 972.00 & 978.00 & 780.00 & 878.00 & 902.00 \\
 &  &  & $\Delta$ & -17.42\% & -21.57\% & -14.57\% & -18.02\% & -18.15\% \\
\cmidrule(lr){3-9}
 &  & \multirow{3}{*}{Overall} & Baseline & 1179.50 & 1023.50 & 583.00 & 927.00 & 928.25 \\
 &  &  & Ours & 992.00 & 819.00 & 494.00 & 774.00 & 769.75 \\
 &  &  & $\Delta$ & -15.90\% & -19.98\% & -15.27\% & -16.50\% & -17.08\% \\
\midrule
\multirow{18}{*}{\textbf{GRPO}} & \multirow{9}{*}{\textbf{Qwen3-1.7b-Base}} & \multirow{3}{*}{Correct} & Baseline & 728.50 & 622.00 & 368.00 & 541.00 & 564.88 \\
 &  &  & Ours & 1071.00 & 681.00 & 417.00 & 606.00 & 693.75 \\
 &  &  & $\Delta$ & +47.01\% & +9.49\% & +13.32\% & +12.01\% & +22.81\% \\
\cmidrule(lr){3-9}
 &  & \multirow{3}{*}{Incorrect} & Baseline & 767.00 & 761.00 & 610.00 & 686.00 & 706.00 \\
 &  &  & Ours & 969.00 & 966.50 & 737.00 & 848.00 & 880.12 \\
 &  &  & $\Delta$ & +26.34\% & +27.00\% & +20.82\% & +23.62\% & +24.66\% \\
\cmidrule(lr){3-9}
 &  & \multirow{3}{*}{Overall} & Baseline & 767.00 & 698.50 & 442.00 & 638.00 & 636.38 \\
 &  &  & Ours & 975.00 & 868.00 & 502.00 & 767.00 & 778.00 \\
 &  &  & $\Delta$ & +27.12\% & +24.27\% & +13.57\% & +20.22\% & +22.25\% \\
\cmidrule(lr){2-9}
 & \multirow{9}{*}{\textbf{Qwen3-4b-Base}} & \multirow{3}{*}{Correct} & Baseline & 1099.50 & 758.00 & 456.00 & 658.50 & 743.00 \\
 &  &  & Ours & 955.00 & 705.50 & 410.00 & 610.00 & 670.12 \\
 &  &  & $\Delta$ & -13.14\% & -6.93\% & -10.09\% & -7.37\% & -9.81\% \\
\cmidrule(lr){3-9}
 &  & \multirow{3}{*}{Incorrect} & Baseline & 945.00 & 1034.00 & 799.50 & 870.50 & 912.25 \\
 &  &  & Ours & 874.00 & 887.00 & 734.00 & 798.00 & 823.25 \\
 &  &  & $\Delta$ & -7.51\% & -14.22\% & -8.19\% & -8.33\% & -9.76\% \\
\cmidrule(lr){3-9}
 &  & \multirow{3}{*}{Overall} & Baseline & 969.00 & 872.50 & 516.50 & 776.00 & 783.50 \\
 &  &  & Ours & 898.50 & 795.50 & 473.00 & 718.00 & 721.25 \\
 &  &  & $\Delta$ & -7.28\% & -8.83\% & -8.42\% & -7.47\% & -7.95\% \\
\midrule
\multirow{9}{*}{\textbf{GSPO}} & \multirow{9}{*}{\textbf{Qwen3-1.7b-Base}} & \multirow{3}{*}{Correct} & Baseline & 796.00 & 680.00 & 445.00 & 615.00 & 634.00 \\
 &  &  & Ours & 863.00 & 601.50 & 389.00 & 557.00 & 602.62 \\
 &  &  & $\Delta$ & +8.42\% & -11.54\% & -12.58\% & -9.43\% & -4.95\% \\
\cmidrule(lr){3-9}
 &  & \multirow{3}{*}{Incorrect} & Baseline & 822.00 & 785.00 & 699.00 & 805.00 & 777.75 \\
 &  &  & Ours & 921.00 & 922.00 & 702.00 & 823.00 & 842.00 \\
 &  &  & $\Delta$ & +12.04\% & +17.45\% & +0.43\% & +2.24\% & +8.26\% \\
\cmidrule(lr){3-9}
 &  & \multirow{3}{*}{Overall} & Baseline & 819.50 & 741.00 & 530.00 & 760.00 & 712.62 \\
 &  &  & Ours & 911.50 & 776.50 & 473.00 & 736.00 & 724.25 \\
 &  &  & $\Delta$ & +11.23\% & +4.79\% & -10.75\% & -3.16\% & +1.63\% \\
\bottomrule
\end{tabular}
}
\caption{Median response length across benchmarks. $\Delta$ denotes the relative difference of ours compared with the baseline.}
\label{tab:response_length_median_all}
\end{table*}

\section{Algorithm Process}\label{APP:Alg}
\begin{algorithm*}[t]
\caption{APRM-enhanced reinforcement learning}
\label{alg:aprm_rl}
\begin{algorithmic}[1]
\Require Initial policy $\pi_{\theta}$, APRM $\mathcal{S}_{\phi}$, outcome verifier $V$, APRM coefficient $\beta$, update interval $U$
\For{RL step $=1,2,\dots$}
    \State Sample a batch of questions $\{x\}$
    \For{each question $x$}
        \State Generate a rollout group $\{y_i\}_{i=1}^{G}\sim \pi_{\theta}(\cdot\mid x)$
        \State Compute outcome rewards $r_i^{\mathrm{out}}=V(x,y_i)$
        \State Compute the base token-level advantages $A_{i,t}^{\mathrm{base}}$ using the underlying RL algorithm
        \State Extract hidden states and attention summaries $(\mathbf{H}_i,\mathbf{A}_i)$ from $\pi_{\theta}$
        \State Compute APRM scores $s_i^{\mathrm{APRM}}=\mathcal{S}_{\phi}(\mathbf{H}_i,\mathbf{A}_i)$
        \State Mean-center APRM scores within the rollout group:
        \[
        \Delta_i^{\mathrm{APRM}}
        =
        s_i^{\mathrm{APRM}}
        -
        \frac{1}{G}\sum_{j=1}^{G}s_j^{\mathrm{APRM}}
        \]
        \State Set $\mathbb{I}_{\mathrm{active}}(x)=1$ if the outcome rewards in the group are not all identical, otherwise $0$
        \State Form the APRM-enhanced advantage:
        \[
        A_{i,t}
        =
        A_{i,t}^{\mathrm{base}}
        +
        \beta\,
        \mathbb{I}_{\mathrm{active}}(x)
        \Delta_i^{\mathrm{APRM}}
        \]
    \EndFor
    \State Update $\pi_{\theta}$ with the original RL objective using $A_{i,t}$ in place of $A_{i,t}^{\mathrm{base}}$
    \If{step is divisible by $U$}
        \State Generate fresh rollouts with the latest policy
        \State Update APRM $\mathcal{S}_{\phi}$ using ranking and counterfactual objectives
    \EndIf
\EndFor
\end{algorithmic}
\end{algorithm*}

This section summarizes the training procedure of APRM-enhanced reinforcement learning. APRM is used as a plug-in internal reward module. The overall process of our algorithm is shown in Algorithm \ref{alg:aprm_rl}.

\section{Proof of Propositions}\label{APP:Proof}
\subsection{Proof of Proposition 1}

Consider a causal self-attention matrix $\mathbf{A}\in\mathbb{R}^{T\times T}$ and a value matrix $\mathbf{V}\in\mathbb{R}^{T\times d_v}$, where $T$ is the sequence length and $d_v$ is the value dimension. The contextual representation is
\[
\mathbf{C}=\mathbf{A}\mathbf{V}.
\]
For standard causal softmax attention,
\[
A_{ij}=0 \quad (j>i), \qquad
A_{ij}>0 \quad (j\leq i), \qquad
\mathbf{A}\mathbf{1}=\mathbf{1}.
\]

Assume that $\operatorname{rank}([\mathbf{V},\mathbf{1}])<T$. By the rank--nullity theorem, there exists a nonzero vector $\mathbf{u}\in\mathbb{R}^{T}$ satisfying
\[
\mathbf{u}^{\top}\mathbf{V}=\mathbf{0},
\qquad
\mathbf{u}^{\top}\mathbf{1}=0.
\]
Let $\mathbf{e}_T$ denote the $T$-th standard basis vector and define
\[
\mathbf{\Delta}=\mathbf{e}_T\mathbf{u}^{\top},
\qquad
\mathbf{A}'=\mathbf{A}+\epsilon\mathbf{\Delta},
\]
where
\[
0<|\epsilon|<
\min_{j:u_j\neq 0}\frac{A_{Tj}}{|u_j|}.
\]
This bound is strictly positive because $A_{Tj}>0$ for all $j\leq T$.

The perturbation modifies only the last row, whose causal support includes all positions. Thus, $\mathbf{A}'$ preserves causal support. Moreover, for $u_j\neq 0$,
\[
A'_{Tj}=A_{Tj}+\epsilon u_j
\geq A_{Tj}-|\epsilon||u_j|>0,
\]
while all other entries remain unchanged. Row normalization is also preserved:
\[
\mathbf{A}'\mathbf{1}
=\mathbf{A}\mathbf{1}
+\epsilon\mathbf{e}_T\mathbf{u}^{\top}\mathbf{1}
=\mathbf{1}.
\]
Hence, $\mathbf{A}'$ is a valid causal attention matrix.

Finally,
\[
\mathbf{A}'\mathbf{V}
=\mathbf{A}\mathbf{V}
+\epsilon\mathbf{e}_T\mathbf{u}^{\top}\mathbf{V}
=\mathbf{A}\mathbf{V}.
\]
Since $\epsilon\neq 0$ and $\mathbf{u}\neq\mathbf{0}$, we have $\mathbf{A}'\neq\mathbf{A}$. Therefore, under the stated rank condition, distinct valid causal attention matrices produce identical contextual representations, establishing that the mapping $\mathbf{A}\mapsto\mathbf{A}\mathbf{V}$ is non-injective over the space of valid causal attention matrices.

\subsection{Proof of Proposition 2}
Under log-loss, the Bayes-optimal predictor outputs the true posterior probability, and its expected risk is bounded below by the conditional entropy.
Let \(Q\) denote process quality, \(B\) denote existing reward signals, and \(Z=(\mathbf{H},\mathbf{A})\) denote APRM internal states.

Under log-loss, the Bayes-optimal predictor using only \((X,B)\) achieves risk
\[
\mathcal{L}_{\mathrm{base}}^*
=
H(Q\mid X,B).
\]

If the predictor additionally conditions on \(Z\), the optimal risk becomes
\[
\mathcal{L}_{\mathrm{APRM}}^*
=
H(Q\mid X,B,Z).
\]

Therefore, the improvement obtained by conditioning on APRM internal states is
\[
\begin{aligned}
\Delta_{\mathrm{IG}}
&=
\mathcal{L}_{\mathrm{base}}^*
-
\mathcal{L}_{\mathrm{APRM}}^* \\
&=
H(Q\mid X,B)-H(Q\mid X,B,Z).
\end{aligned}
\]

By the definition of conditional mutual information,
\[
I(Q;Z\mid X,B)
=
H(Q\mid X,B)-H(Q\mid X,B,Z).
\]
Thus,
\[
\Delta_{\mathrm{IG}}
=
I(Q;Z\mid X,B)\ge 0.
\]

Moreover,
\[
I(Q;Z\mid X,B)>0
\]
if and only if
\[
Q \not\!\perp\!\!\!\perp Z \mid X,B.
\]
The internal attention structures remain predictive of process quality after conditioning on existing reward signals. Specifically, in the next subsection \ref{app:info_gain}, we experimentally demonstrate why this condition holds.

\subsection{Additional Evidence for Information Gain}
\label{app:info_gain}

We provide a supplementary residual information test for Proposition~\ref{proposition 2}.
We evaluate two independently sampled rollout sets, generated by
Qwen3-4b-Base and Qwen3-1.7b-Base, respectively.

For each dataset, we regress the LLM-judged process quality \(Q\) on existing
reward signals \(B\), including correctness and log response length, together
with prompt fixed effects \(X\). We then test whether adding the APRM score
\(Z\) improves the prediction:
\[
Q \sim X+B,
\qquad
Q \sim X+B+Z .
\]

\begin{figure*}[h]
    \centering
    \includegraphics[width=1\linewidth]{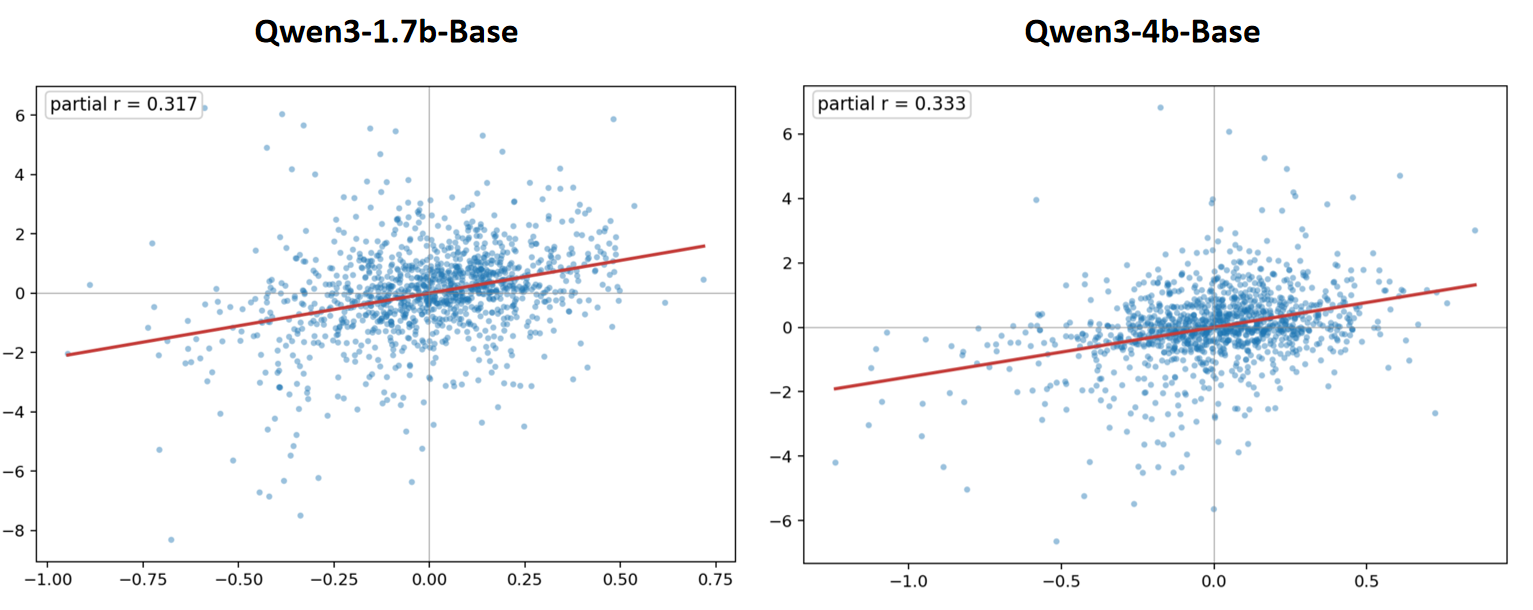}
    \caption{Residual information test for APRM. The two panels show the residual correlation between APRM score and LLM-judged process quality after controlling for correctness, response length, and prompt fixed effects.}
    \label{fig:info_gain}
\end{figure*}
As shown in Figure \ref{fig:info_gain}, the consistent positive residual trends in both rollout sets show that APRM retains predictive signal for process quality after removing correctness, response length, and prompt-specific effects, thereby empirically supporting Proposition~\ref{proposition 2} that \(I(Q;Z\mid X,B)>0\).

\end{document}